\documentclass[final,3p,times,onecolumn]{elsarticle}

\usepackage{amssymb}
\usepackage{multicol}        
\usepackage[bottom]{footmisc}
\usepackage{amsmath,epsfig}
\usepackage{algorithm}
\usepackage{algorithmic}
\usepackage{soul}
\usepackage{color}
\usepackage{setspace}

\long\def\comment#1{}

\journal{Journal of Visual Communication and Image Representation}

\begin{document}

\begin{frontmatter}

\title{Statistical Denoising for single molecule fluorescence microscopic images}

\author[label1]{Ji Won Yoon} 
\address[label1]{Korea University, Korea, jiwon\_yoon@korea.ac.kr}

\begin{abstract}

Single molecule fluorescence microscopy is a powerful technique for uncovering detailed information about biological systems, both {\it in vitro} and {\it in vivo}. In such experiments, the inherently low signal to noise ratios  mean that accurate algorithms to separate true signal and background noise are essential to generate meaningful results. To this end, we have developed a new and robust method to reduce noise in single molecule fluorescence images by using a Gaussian Markov Random Field (GMRF) prior in a Bayesian framework. Two different strategies are proposed to build the prior - an intrinsic GMRF, with a stationary relationship between pixels and a heterogeneous intrinsic GMRF, with a differently weighted relationship between pixels classified as molecules and background. Testing with synthetic and real experimental fluorescence images demonstrates that the heterogeneous intrinsic GMRF is superior to other conventional de-noising approaches.

\end{abstract}


\end{frontmatter}

\section{Introduction}
\label{section: introduction}
Using fluorescence microscopy with single-molecule sensitivity, it is now possible to track the movement of individual fluorophore-tagged molecules such as proteins and lipids in the cell membrane with nanometer precision. This ability has been used to characterize the diffusional properties of molecules \cite{Saxton97:SingleParticleTracking}, as well as monitor their organization relative to other molecules \cite{Dunne09:SMF} and cellular structures \cite{Andrews08:SingleParticleTracking}, with the potential to work below the diffraction limit of light. 

The design of automatic, efficient computer algorithms for data analysis is an extremely important facet of such experiments. Ideally, the program should be able to tolerate the low signal to noise ratios inherent to single-molecule data, must work at high object density and be able to cope with large data volumes at reasonable speed. Suitable methods can usually be decomposed into several sequential steps: filtering of the data, identification of fluorescent objects and their precise positions, and ‘tracking’ of these objects by linking together their positions over an image sequence. For precise tracking care must be taken to optimise each of these stages.

Previous work in Single Molecule Fluorescence image processing has focussed on the tracking step of the problem, with several suggested approaches resulting in excellent performance, including a grid algorithm \cite{Bruckbaur07:SMF}, and several Monte Carlo-based algorithms \cite{Yoon08:SMF,Dunne09:SMF}. However, with one exception \cite{Yoon08:SMF}, the initial de-noising step utilized conventional de-noising algorithms, such as Gaussian filtering. Afterwards, the detection algorithms are applied to the de-noised images in order to separate signals and background noise. As a result, the tracking algorithms had to assume a cluttered environment with false alarms and missed detections, impacting their performance. Thus in this paper we concentrate on the first step in the data analysis procedure: ‘data filtering’, or ‘de-noising’.

De-noising is commonly carried out using one of two approaches: Gaussian smoothing or wavelet de-noising. In the first, the image is smoothed through convolution with a Gaussian or average kernel, which is related to matched filtering for additive uncorrelated noise \cite{Turin60:MatchedFilter}. Use of a Gaussian filter is justified since the fluorescence profile of an individual molecule is well approximated by (and hence frequently fitted to) a Gaussian function \cite{Zhang07:GA4Fluorescence}. In cases where the noise addition is linear or stationary, a Wiener filter is also used to reduce the noise of the image \cite{Wiener64:WienerFilter}. For the non-stationary problem, wavelet based de-noising algorithms are also well-known for image restoration \cite{Donoho93:nonlinearwavelet,Marin02:SpotDetection}.

In this work we develop and describe a new de-noising algorithm based on a Gaussian Markov Random Field (GMRF) model. Our proposed algorithm is a fully Bayesian approach with few tuning parameters: we need to set only a small relative threshold parameter and specify hyper-parameters to build prior distributions. The performance of this method relative to previous methods is evaluated using both synthetic and real single molecule data and found to display significant advantages.

The paper is structured as follows. We first introduce the Gaussian Markov random field in a latent Gaussian model (Section \ref{section: Gaussian Models}) which is then used to propose mathematical models for de-noising images (Sections \ref{section: Mathematical Model} and \ref{section: Posterior Calculation}). Finally, we compare the results of the various algorithms when applied to synthetic and real data (Section \ref{section: Simulated Results}).

%

\section{Intrinsic Gaussian Models}
\label{section: Gaussian Models}
Gaussian Markov random fields (GMRFs) are Gaussian fields defined on a discrete grid with a Markov property of conditional independence of a component with all others given its neighbours \cite{Mardia88:GMRF,GMRFbook:Rue}. They have seen widespread application in statistical modelling, for example in spatio-temporal models \cite{Besag91:GMRF} and dynamic linear models \cite{West97:LDM}. 

In this paper, a GMRF ${\bf f}$ is an one dimensional vector which corrsponds to a two dimensional image lattice $\{(i,j) \, | \, i=1,\ldots,n_1; j=1,\ldots,n_2\}$ \cite{GMRFbook:Rue}. Let $\Delta_{i,j}$ be differenced values of ${\bf f}$ at a site $(i,j)$ on the lattice with neighbours. There are many possible ways to define a GMRF for ${\bf f}$ through the $\Delta_{i,j}$. For example, a first order model can be defined through assuming the differences with the 4 nearest neighbours $\eta_{i,j} = \{(i-1,j),(i+1,j),(i,j-1),(i,j+1)\}$,
\[
\Delta_{i, j} = \sum_{(k,l) \in \eta_{i,j}}( f_{k, l} - f_{i, j}),
\]
to be Gaussian with mean 0 and precision (inverse of variance) $\kappa_f$. The distribution of ${\bf f}$ is of multivariate Gaussian form:
\begin{eqnarray}
p({\bf f}|\kappa_f)
&\propto&\exp\left\{-\frac{\kappa_{f}}{2}\sum_{i=1}^{n_1}\sum_{j=1}^{n_2}\Delta_{i, j}^{2}\right\}\cr
&=& \exp\left\{-0.5\kappa_{f}({\bf D}{\bf f})^{T}({\bf D}{\bf f})\right\}\cr
&=& \exp\left\{-0.5\kappa_{f}{\bf f}^{T}{\bf Q}_{f}{\bf f}\right\},
\label{eq: prior for f}
\end{eqnarray}
where the precision matrix (inverse of covariance matrix) ${\bf Q}_{f}={\bf D}^{T}{\bf D}$, and ${\bf D}$ is a $n_1n_2 \times n_1 n_2$ matrix with elements  $D_{m,m}=-\sum_{d=1, d\neq m}^{n_{1}n_{2}}D_{m, d}$, $D_{m_1,m_2}=-1$ if the pixels represented by the $m_1$th and $m_2$th components are neighbours, and 0 otherwise. The resulting precision matrix ${\bf Q}_f$ is not of full rank, in which case it is known as an intrinsic GMRF (IGMRF). This form is used often as a prior in Bayesian inference. For de-noising, it has the desirable properties of placing more prior weight on smooth images while avoiding the specification of a mean value of ${\bf f}$. Under very general conditions, the posterior of ${\bf f}$ will become a proper distribution \cite{GMRFbook:Rue}.

\section{The De-noising Model}
\label{section: Mathematical Model}
Note that we focus on de-noising a single image in this paper although we have a sequence of images for processing. For simplicity, we assume images in the sequence are independent so that  the de-noising algorithm can be applied separately. The dependency of spots across the sequence of images is accounted for in the tracking process.

\subsection{Linear Model for an intrinsic GMRF}
We assume a Gaussian model for an observed image ${\bf y}$ in terms of the underlying signal ${\bf f}$  and some regression coefficients $\gamma=(\gamma_1,\gamma_2,\gamma_3)$:
\begin{equation}
{\bf y} = {\bf z}^{T}\gamma + {\bf f} + \epsilon
\label{eq: linear model for denoising}
\end{equation}
where $\epsilon \sim \mathcal{N}(\cdot; {\bf 0}, {\kappa_{l}}^{-1}{\bf I})$ for an unstructured term and $\mathcal{N}$  denotes the Gaussian distribution. In this model, ${\bf z}$ are lattice location indices so that ${\bf z}^{T}\gamma$ models any global background noise trend. Therefore, the set of parameters are denoted $\theta=\{\kappa_{l}, \kappa_{f}\}$.

\subsection{Priors}
We assume conjugate prior forms of the regression coefficients
\begin{equation}
\gamma \sim p(\gamma)=\mathcal{N}(\gamma; {\bf 0}, {\bf Q}_{\gamma}^{-1}),
\label{eq: prior for regression coefficients}
\end{equation} 
where ${\bf Q}_{\gamma} = \frac{1}{10^{3}}{\bf I}_{3\times 3}$, giving a weak prior, and
\begin{eqnarray}
\kappa_{f} &\sim& \mathcal{G}(\cdot; \alpha_{f}, \beta_{f});\cr
\kappa_{l} &\sim& \mathcal{G}(\cdot; \alpha_{l}, \beta_{l}),
\end{eqnarray}
where $\mathcal{G}(\cdot; \alpha, \beta)$ denotes the gamma distribution with hyper-parameters $\alpha$ and $\beta$.

According to the linearity of the hidden parameters in Eq. (\ref{eq: linear model for denoising}), we can apply Rao-Blackwellization technique \cite{Casella96:Rao-Blackwellisation} to reduce the dimension of the hidden variables and variances. Now, we have marginalized likelihood defined by
\begin{eqnarray}
p({\bf y}|{\bf f}, \theta) &=& \int_{\gamma} p({\bf y}|{\bf f}, \gamma, \theta)p(\gamma|\theta) d \gamma\cr
&=& \int_{\gamma} \mathcal{N}({\bf y}; {\bf z}^{T}\gamma+{\bf f}, \kappa_{l}^{-1}{\bf I})\mathcal{N}(\gamma; {\bf 0}, {\bf Q}_{\gamma}^{-1})d\gamma\cr
&=& \mathcal{N}({\bf y}; {\bf f}, \Phi)
\label{eq: marginalized likelihood}
\end{eqnarray}
where $\Phi = \kappa_{l}^{-1}{\bf I}+{\bf z}^{T}{\bf Q}_{\gamma}^{-1}{\bf z}$.

\subsection{Problems with the IGMRF prior}
\cite{GMRFbook:Rue} describes several limitations of the intrinsic GMRF. The limitations will bring following problems in de-noising images:
\begin{enumerate}
\item Unwanted blurring effects can be introduced because the IGMRF penalises discrete boundaries or sharp gradients between neighbours;
\item Weak signals can be ignored by smoothing with the background;
\item Stationary (homogeneous) fields across an image rarely exist in practice.
\end{enumerate}
In spite of its convenient mathematical properties, these three concerns present difficulties for this application. A variant of the conventional IGMRF is therefore introduced that assigns different interaction weights to pixel neighbours according to whether the pixel is classified as being part of a spot (the molecule of interest is present) or the background. We name this variant of the IGMRF as the Heterogeneous IGMRF (HIGMRF).

\subsection{Heterogeneous IGMRF (HIGMRF)}
\label{section: Heterogeneous GMRF (HGMRF)}
In this application, pixels in the image can be classified as either a spot  or background.  This provides a way to allow for non-stationarity in an IGMRF. For pixel $(i,j)$ let ${\bf e}_{i,j} = 1$ if it is a spot and  ${\bf e}_{i,j}=0$ if it is background.  The conditional distribution  $p({\bf f} \, | \, {\bf e})$ is of the same Gaussian form of Eq.\  (\ref{eq: prior for f}), and the precision matrix ${\bf Q}_f$ is defined through differences $\Delta_{i,j}$ as with the IGMRF, but the differences are weighted by $\lambda$ between a background pixel and neighbouring spot pixels, so
\begin{equation}
\Delta_{i,j} = \begin{cases}
\sum_{(k,l) \in \eta_{i,j}} (f_{k,l}-f_{i,j}), & \mbox{if } {\bf e}_{i,j}=1, \\
\sum_{(k,l) \in \eta_{i,j}} \lambda^{1-e_{k,l}} (f_{k,l}-f_{i,j}), & \mbox{if } {\bf e}_{i,j}=0,
\end{cases}
\end{equation}
for $\lambda > 1$.  The effect of this prior is to give higher probability to images with large contiguous background regions whilst preventing over-smoothing of spots. The value of $\lambda$ must be specified; $\lambda=50$ is used here and works well for the images used in this paper. This model is related to the Ising model commonly used in image processing \cite{Gelman84:GibbsSampling}. (Although the ${\bf e}$ can be estimated from the $p({\bf e}|{\bf f})$, we do not estimate ${\bf e}$ in Gibbs scheme because it is time consuming.

\section{Posterior Calculations}
\label{section: Posterior Calculation}
The unknown parameters in the marginalized form are ${\bf f}$ and $\theta = \{\kappa_{l}, \kappa_{f}\}$ and so the goal is to evaluate $p({\bf f}, \theta \, | \, {\bf y})$. This is implemented by a Gibbs sampler that samples ${\bf f}$ from
\begin{equation}
p({\bf f}|{\bf y}, \theta)  \propto p({\bf y}|{\bf f}, \theta) p({\bf f}|\theta)  = \mathcal{N}({\bf f}; \mu^{*}, \Sigma^{*})
\end{equation}
where $\Sigma^{*} = (\Phi^{-1}+\kappa_{f}{\bf Q}_{f})^{-1}$ and $\mu^{*} = [{\bf I}+\kappa_{f}\Phi{\bf Q}_{f}]^{-1}{\bf y}.$

Then $\kappa_{l}$ and $\kappa_{f}$  are  sampled separately from $p(\theta | {\bf f}, {\bf y})$. However, it is not straightforward to draw samples from the marginalized conditional posterior since there is no exact conditional distribution which we can easily generate samples. Thus, we draw the $\gamma$ from the complete posterior (which is a sort of Data Augmentation technique) and then we generate samples on the model parameters $\kappa_{l}$ and $\kappa_{f}$. 

\begin{equation}
\gamma \sim p(\gamma|{\bf y}, {\bf f}, \theta)=\mathcal{N}(\gamma; {\bf m}, {\bf C})
\end{equation}
where ${\bf C}= (\kappa_{l}{\bf z}{\bf z}^{T}+{\bf Q}_{\gamma})^{-1}$ and ${\bf m}=\kappa_{l}{\bf C}{\bf z}({\bf y}-{\bf f})$.

Now, we can generate samples of $\kappa_{l}$ and $\kappa_{f}$ separately from $p(\theta |{\bf y},  {\bf f}, \gamma)$, which are of the form:
\begin{eqnarray}
\kappa_{l} &\sim& p(\kappa_{l}|{\bf y}, {\bf f}, \gamma)= \mathcal{G}\left(\cdot; \alpha_{l}^{*}, \beta_{l}^{*}\right)\cr
\kappa_{f} &\sim& p(\kappa_{f}|{\bf f}) = \mathcal{G}(\cdot; \alpha_{f}^{*}, \beta_{f}^{*})
\label{eq: calculate kappa}
\end{eqnarray}
where 
\begin{eqnarray}
\alpha_{l}^{*} &=&  N/2+\alpha_{l}\cr
\beta_{l}^{*} &=& \left[\frac{({\bf y}-{\bf z}^{T}\gamma -{\bf f})^{T}({\bf y}-{\bf z}^{T}\gamma -{\bf f})}{2}+\frac{1}{\beta_{l}}\right]^{-1}\cr
\alpha_{f}^{*} &=& N/2+\alpha_{f}\cr
\beta_{f}^{*} &=& \left[\frac{f^{T}{\bf Q}_{f}f}{2}+\frac{1}{\beta_{f}}\right]^{-1}
\label{eq: alphas and betas for kappas}
\end{eqnarray}

\subsection{Sampling ${\bf f}$ in the HIGMRF case}
When ${\bf f}$ is an HIGMRF, it depends on ${\bf e}$.  Good performance of the algorithm can be obtained if ${\bf e}$ is updated at each iteration of the Gibb's sampler by a locally defined threshold $h_{local}$; ${\bf e}_{ij}=0$ if ${\bf f}_{ij}<h_{local}$ (background) and otherwise ${\bf e}_{ij}=1$ (spot), reflecting the fact that the background is darker. The details of the algorithm are described in algorithm \ref{algorithm: get binary image}. The threshold is specified through a parameter $h$; empirically it has been found that $h=0.1$ works well for the images used in this paper. More details to find such a hard threshold $h$ will be referred to in discussion section.
\begin{algorithm}[h!]
\caption{${\bf e}\leftarrow getBinaryImage({\bf f})$}
\label{algorithm: get binary image}
\begin{algorithmic}[1]
\STATE $h \leftarrow 0.1$
\FOR{$i=1$ to $n_{1}$}
	\FOR{$j=1$ to $n_{2}$}
	    \STATE Create a window ${\bf f}_{win}$ centred at $(i,j)$.
	    \STATE Calculate a local threshold by $h_{local} = \mu_{local}+h \sigma_{local}$ where $\mu_{local}$ and $\sigma_{local}$ are the mean and standard deviation of pixels of ${\bf f}_{win}$ respectively.
	    \IF{${\bf f}_{i, j} >= h_{local}$}
	        \STATE ${\bf e}_{i,j} = 1$
	    \ELSE 
	       \STATE ${\bf e}_{i,j} = 0$    
	    \ENDIF
    \ENDFOR
\ENDFOR
\end{algorithmic}
\end{algorithm}

\subsection{Algorithm for HIGMRF}
\label{section: Algorithm for HGMRF}
The Gibb's sampler is initialized with ${\bf e}=0$, ${\bf Q}_f$ defined as in the usual IGMRF case of Section \ref{section: Gaussian Models} . The $t$th iteration of the Gibb's sampler is:
\begin{itemize}
\item $\gamma^{(t)}$ is sampled from $p(\gamma|{\bf y}, {\bf f}^{(t-1)}, \theta^{(t-1)})$.
\item $\theta^{(t)}=(\kappa_{l}^{(t)}, \kappa_{f}^{(t)})$ is sampled from $p(\theta|{\bf y}, {\bf f}^{(t-1)}, \gamma^{(t)})$.
\item ${\bf f}^{(t)}$ is sampled from the marginalized posterior $p({\bf f}|{\bf y}, \theta^{(t)})$.
\item If the HIGMRF prior is used for ${\bf f}$:
\begin{itemize}
\item ${\bf e}^{(t)}$ is calculated from ${\bf f}^{(t)}$ from algorithm \ref{algorithm: get binary image}.
\item ${\bf Q}_{f}^{(t)}$ is calculated from ${\bf e}^{(t)}$.
\end{itemize}
\end{itemize}
The de-noised image is reported as the sample average of the ${\bf f}^{(t)}$, following an initial burn-in period.

\begin{figure*}[ht!]
\centering
\begin{tabular}{|c|ccccc}
\includegraphics[width=0.9in, height=0.9in]{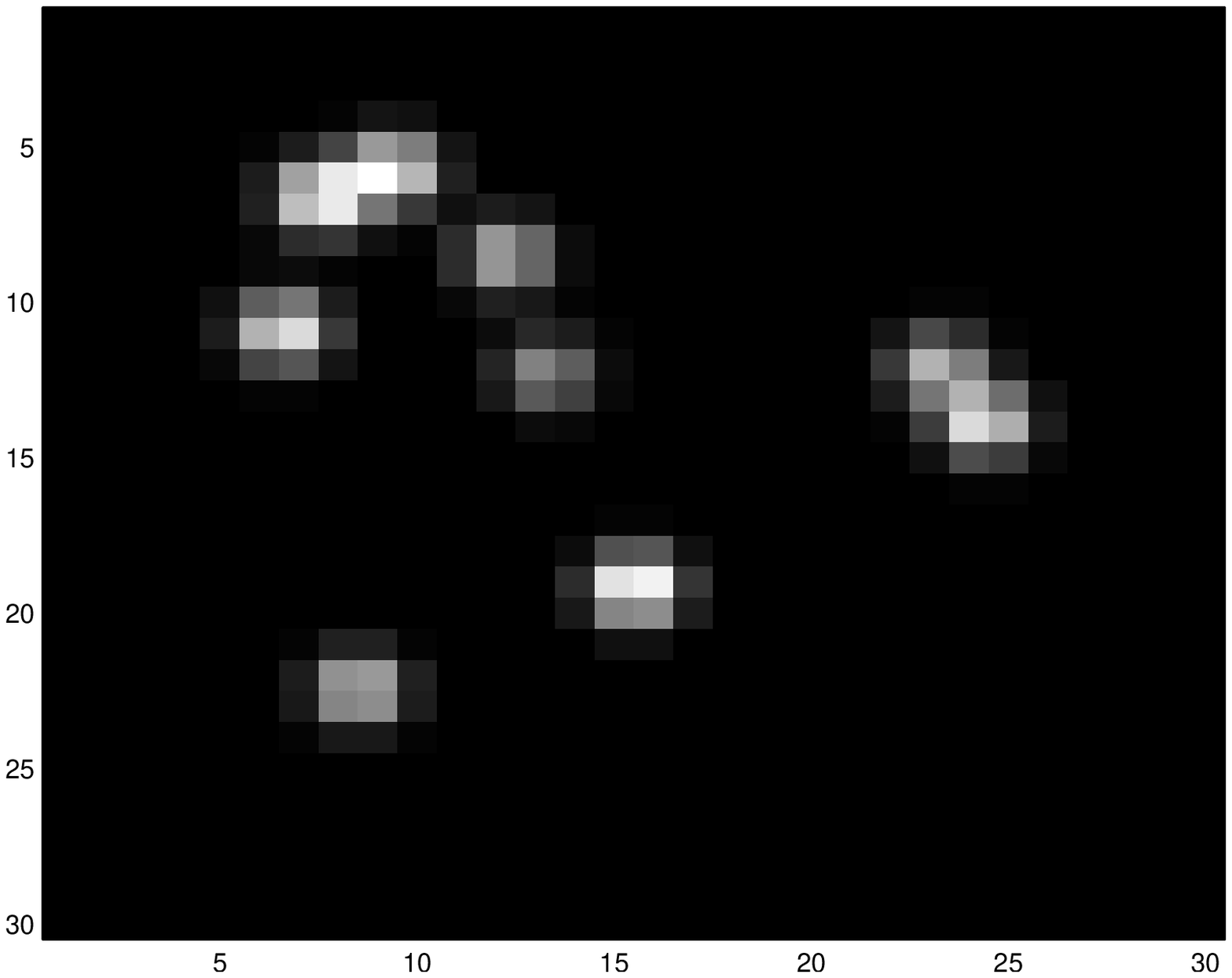}&
\includegraphics[width=0.9in, height=0.9in]{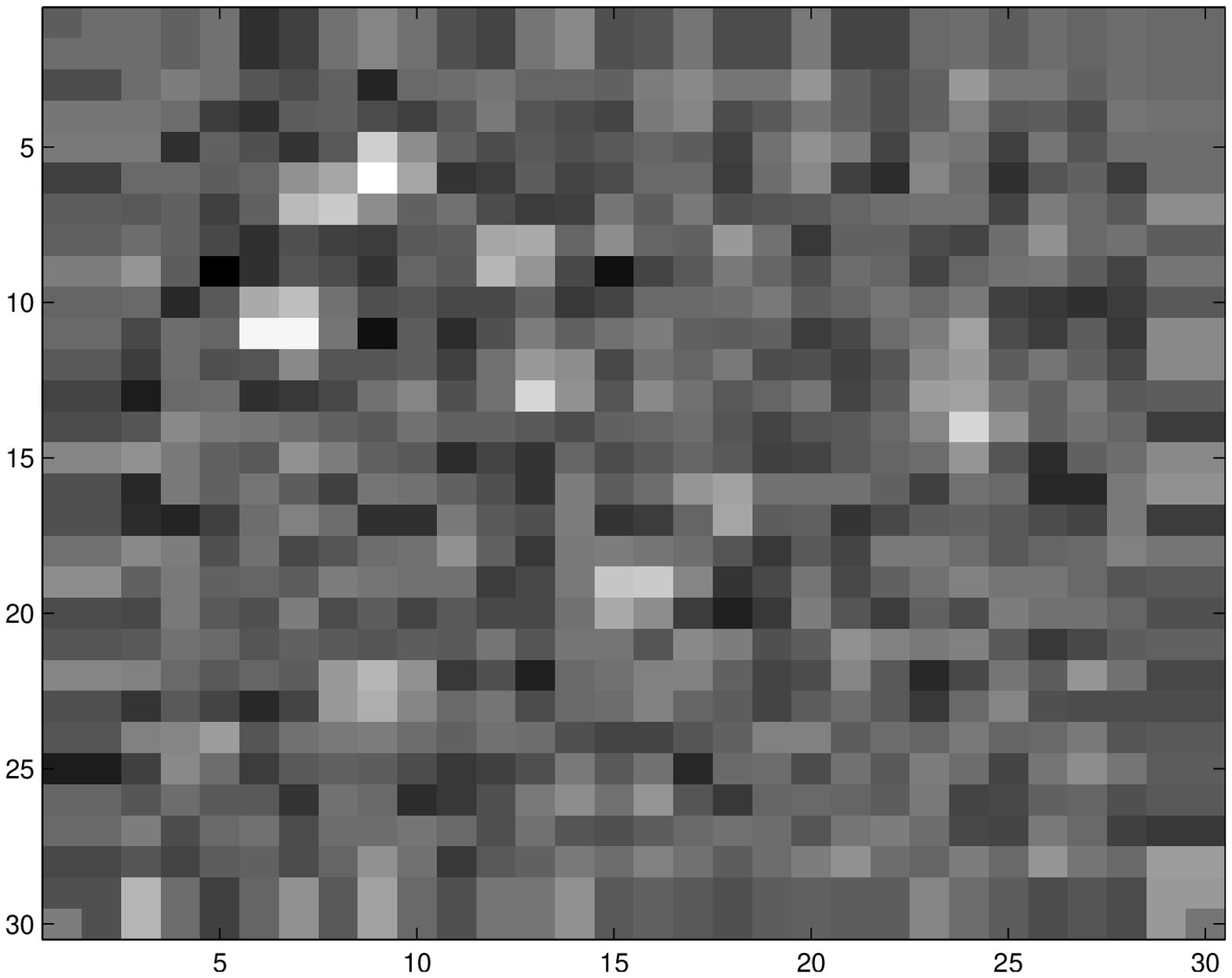}&
\includegraphics[width=0.9in, height=0.9in]{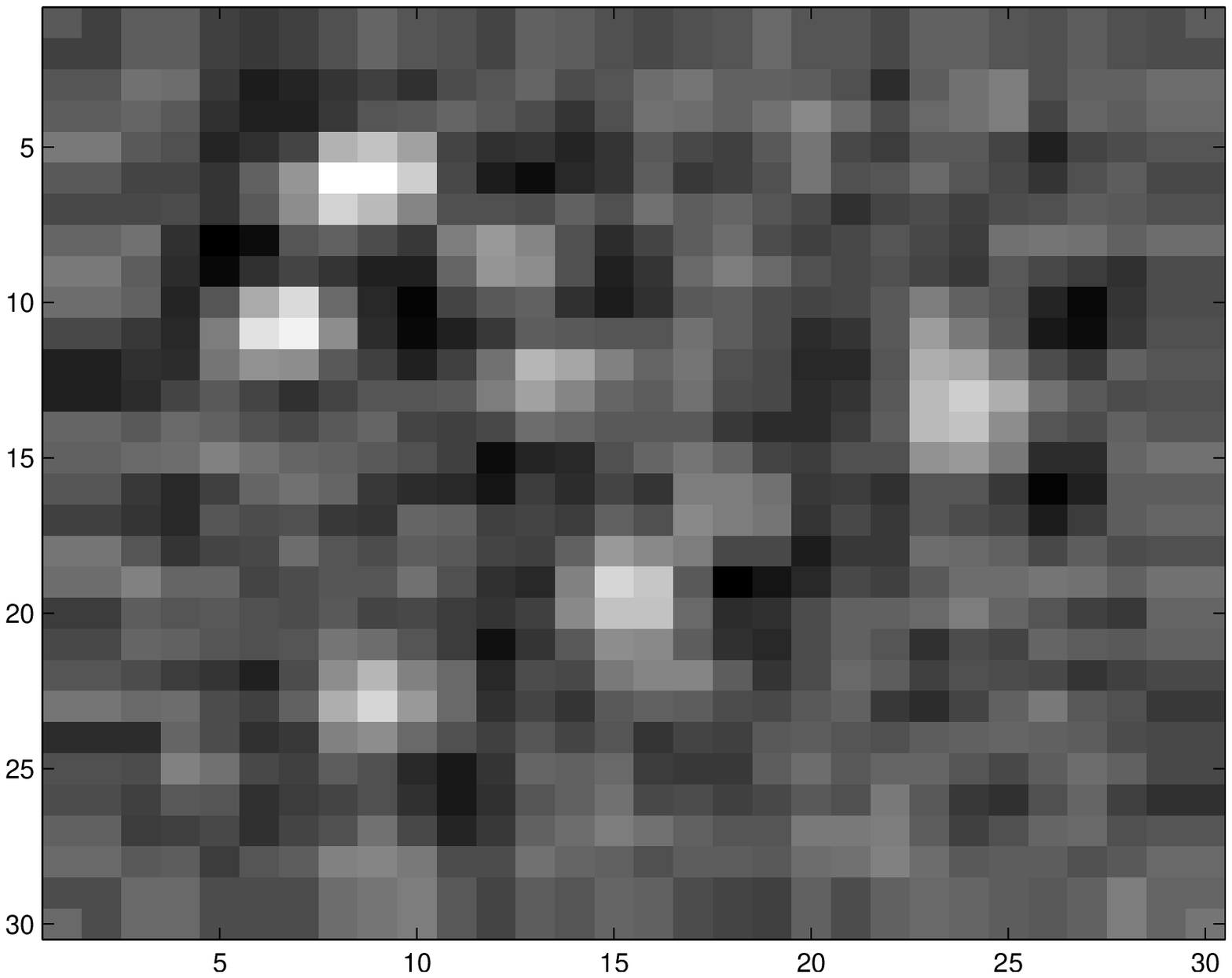}&
\includegraphics[width=0.9in, height=0.9in]{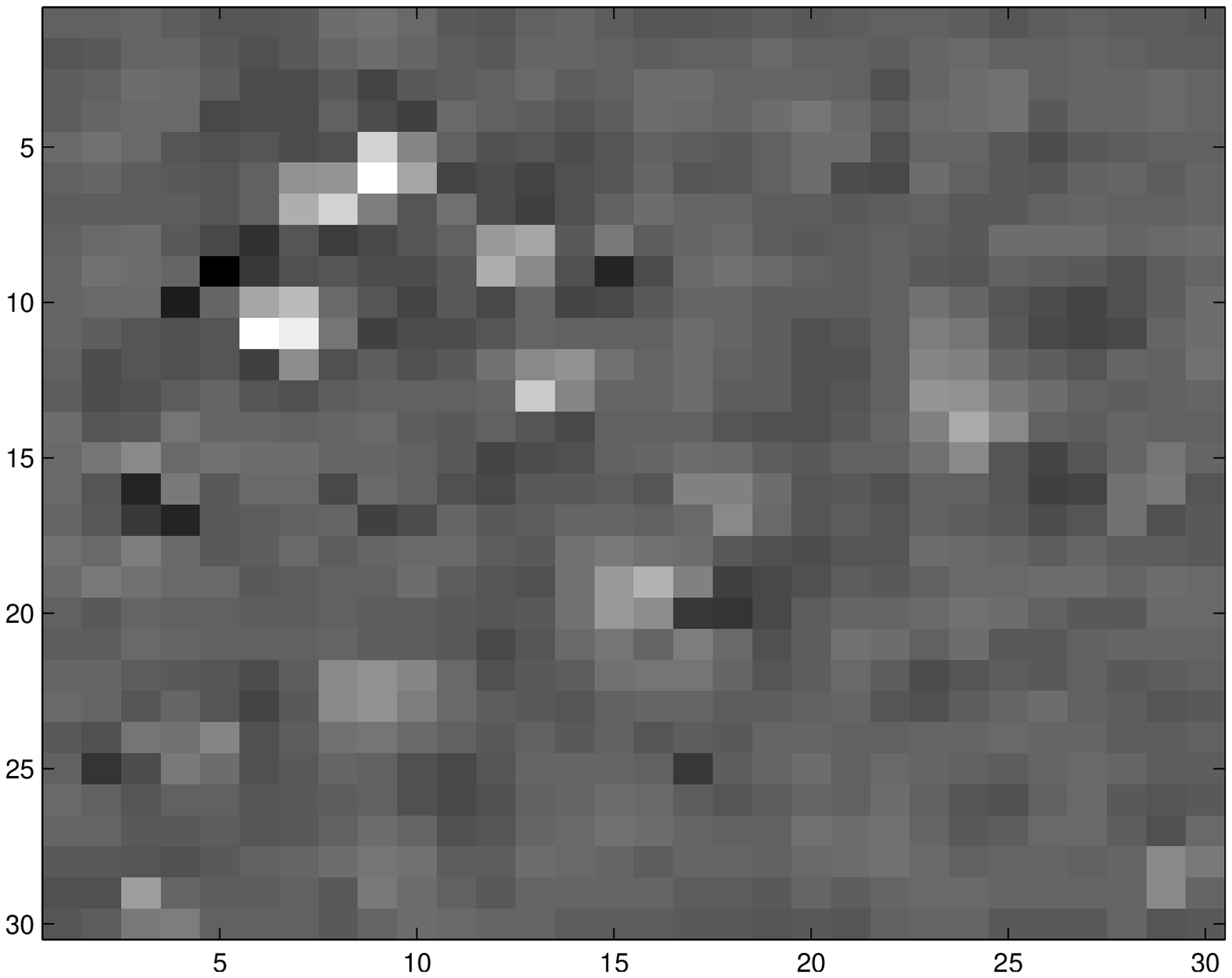}&
\includegraphics[width=0.9in, height=0.9in]{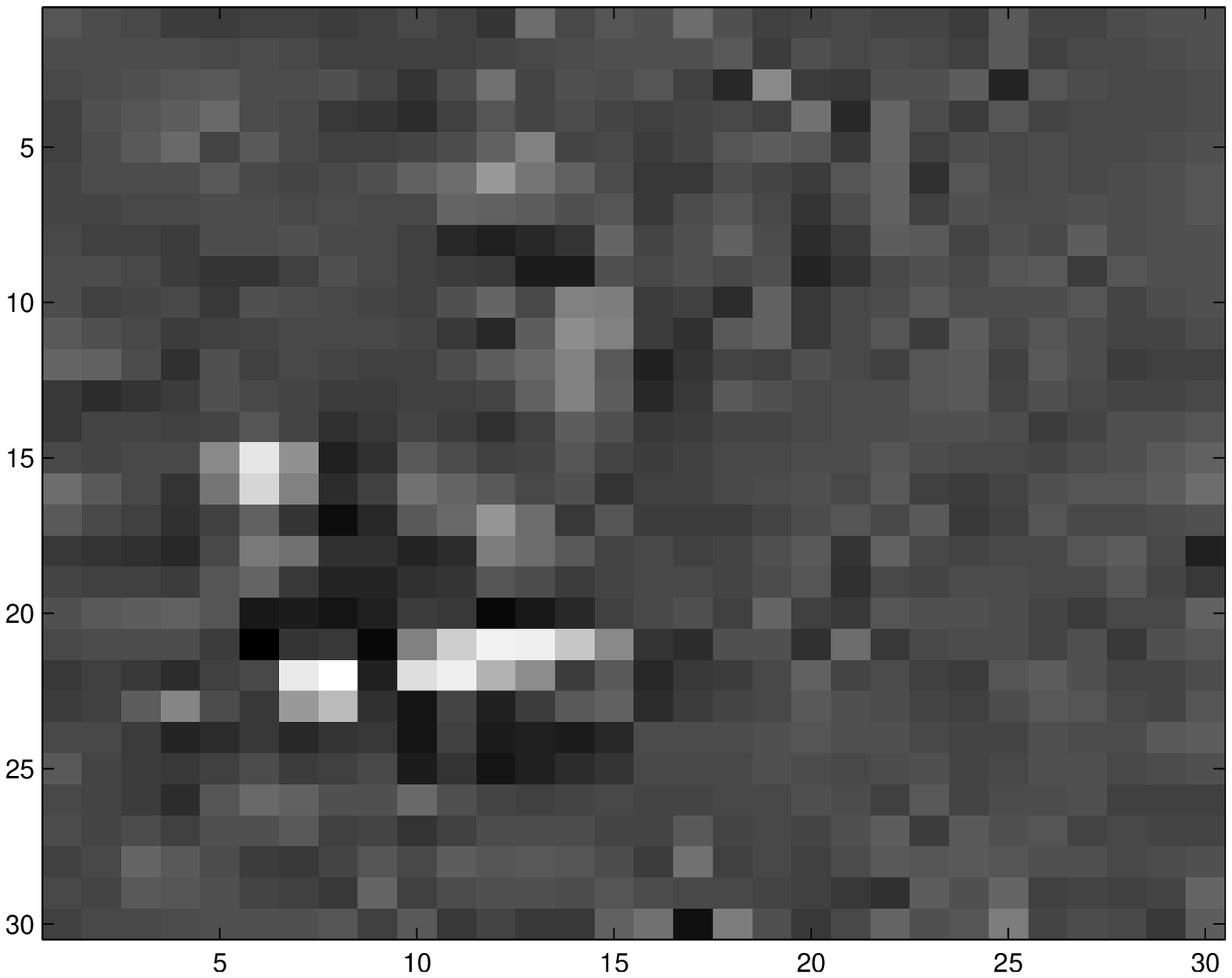}&
\includegraphics[width=0.9in, height=0.9in]{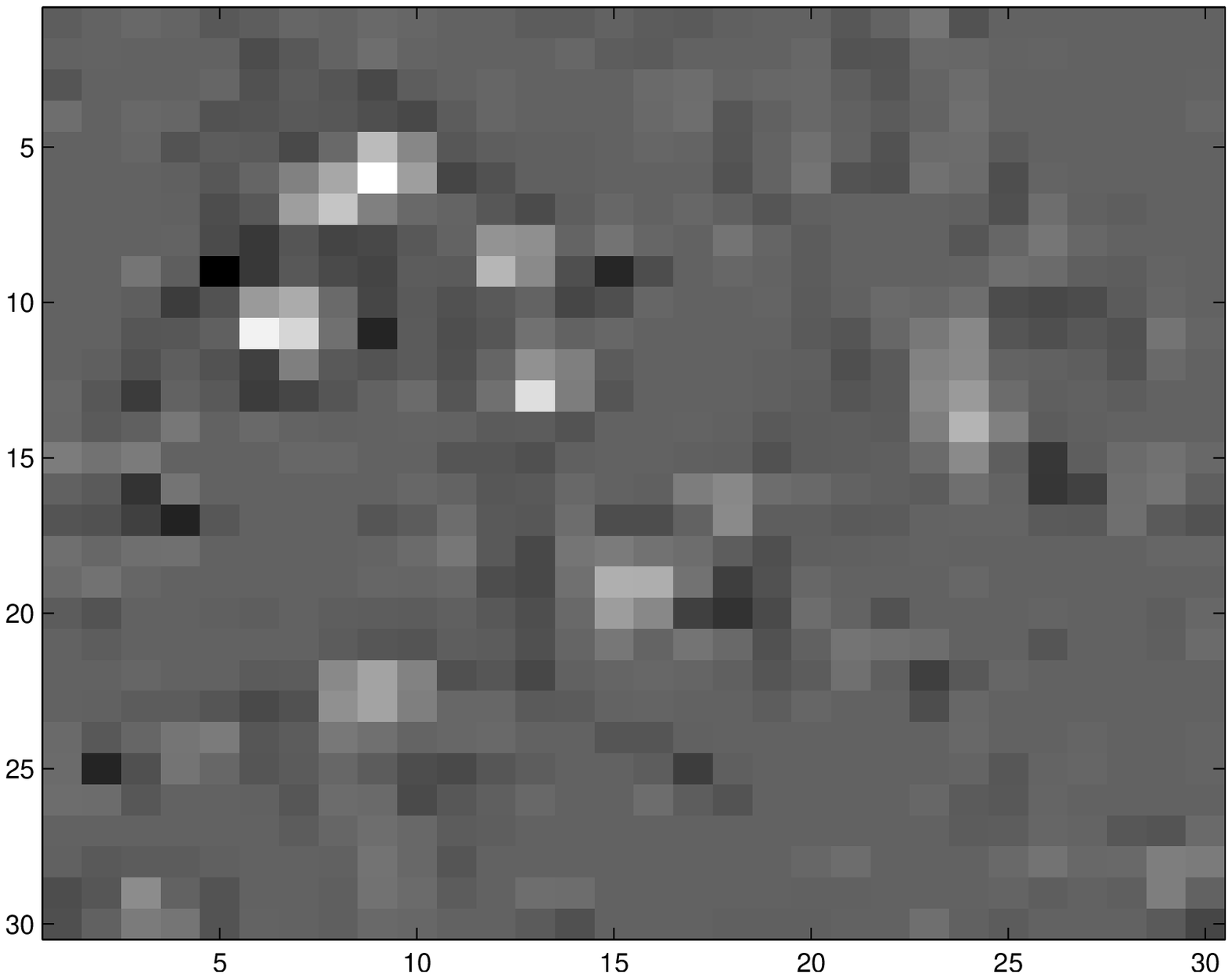}\cr
(a) Ground&
(b) Gaussian & 
(c) Average &
(d) Wiener & 
(e) Wavelet&
(f) LGW \cr
Truth&
(Ga) & 
(Av) &
(Wi) & 
(Wav)&
\cr
\includegraphics[width=0.9in, height=0.9in]{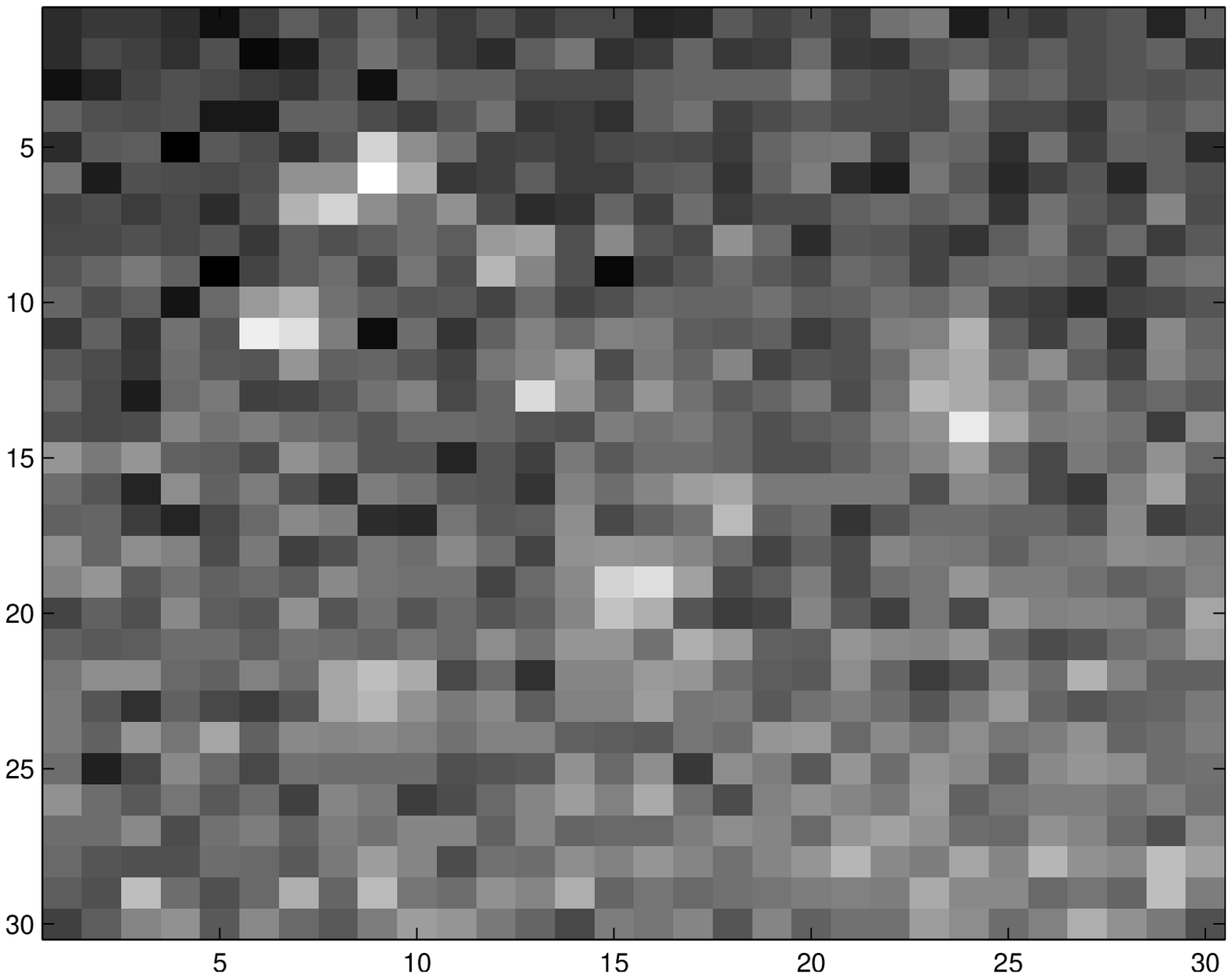}&
\includegraphics[width=0.9in, height=0.9in]{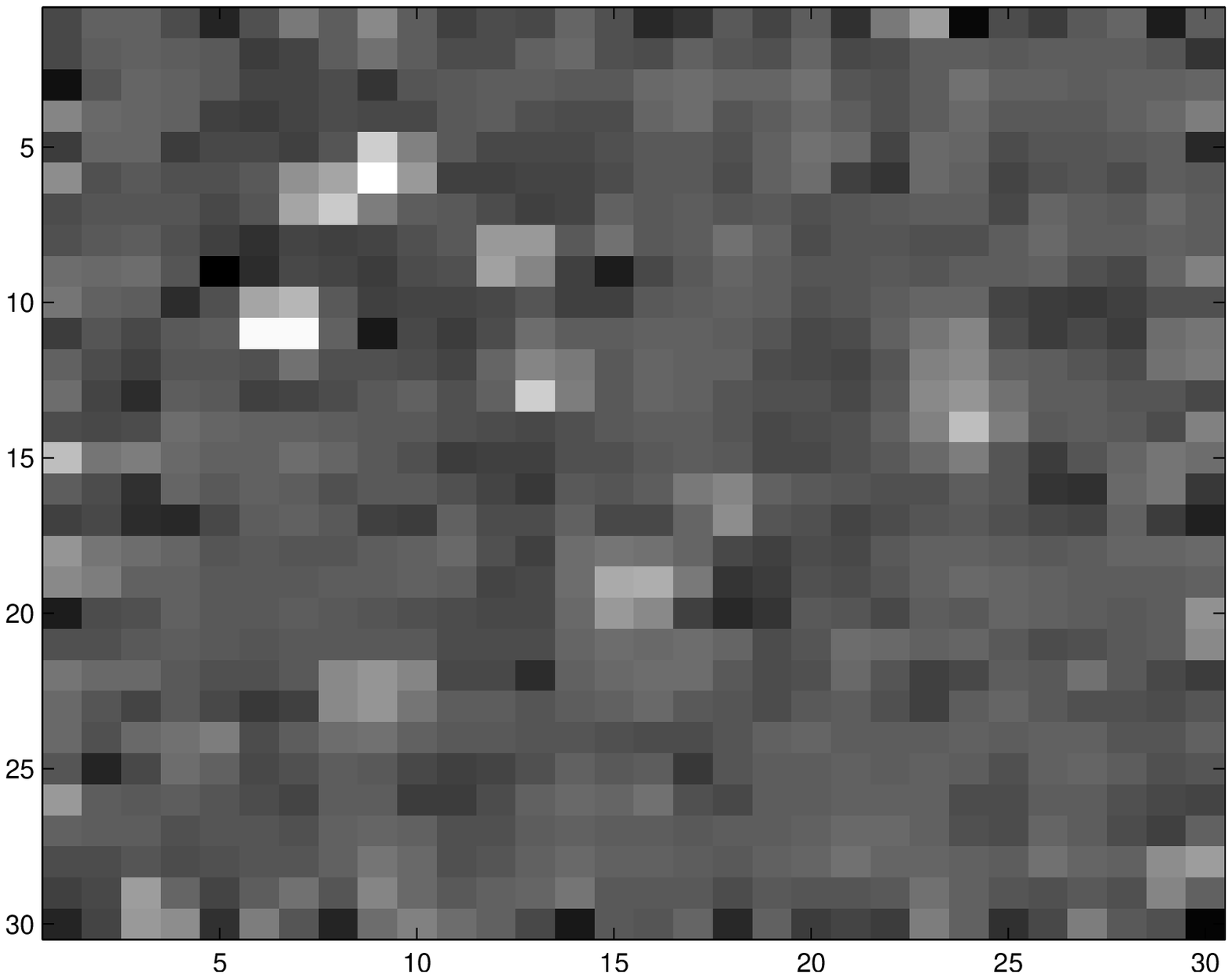}&
\includegraphics[width=0.9in, height=0.9in]{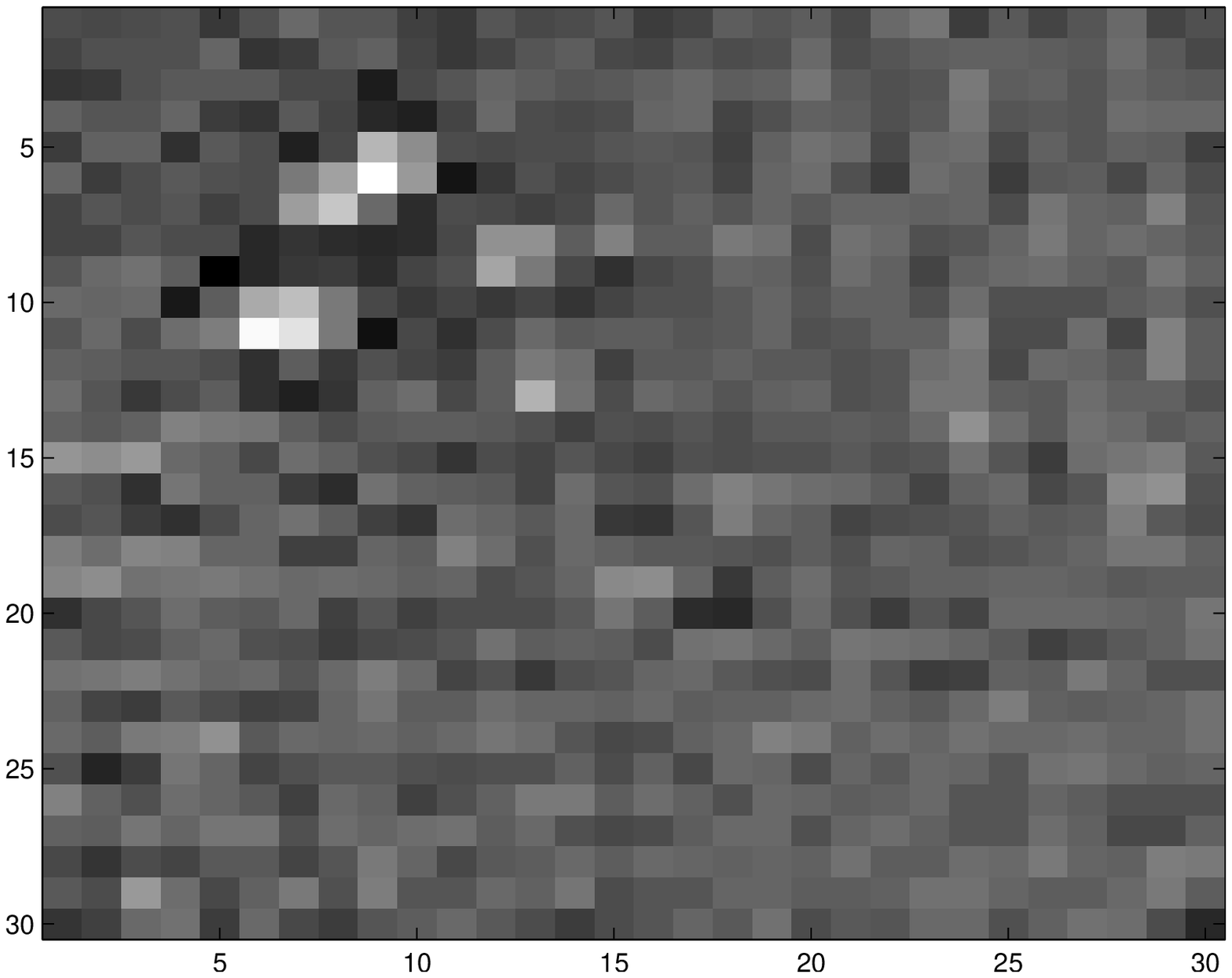}&
\includegraphics[width=0.9in, height=0.9in]{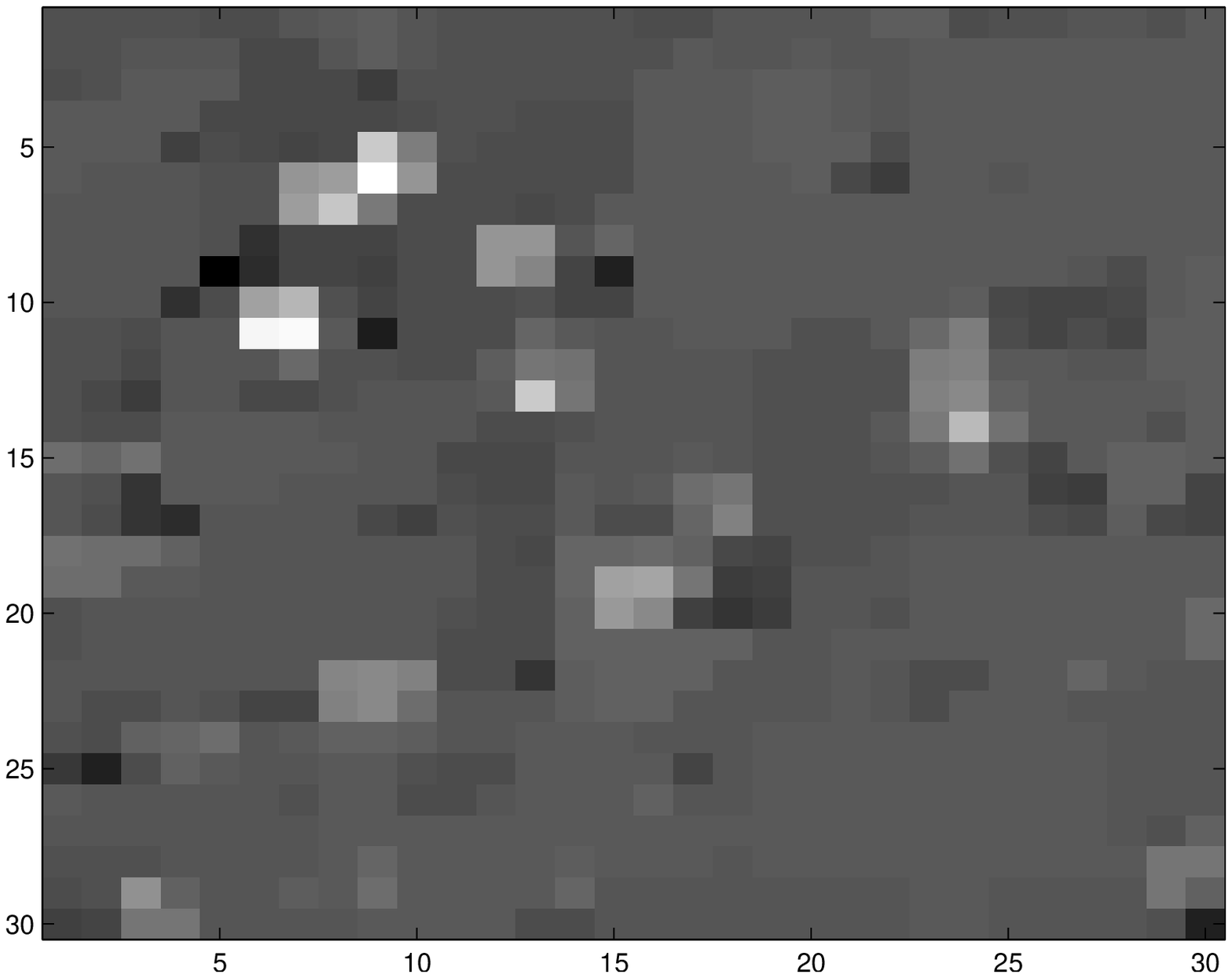}&
\includegraphics[width=0.9in, height=0.9in]{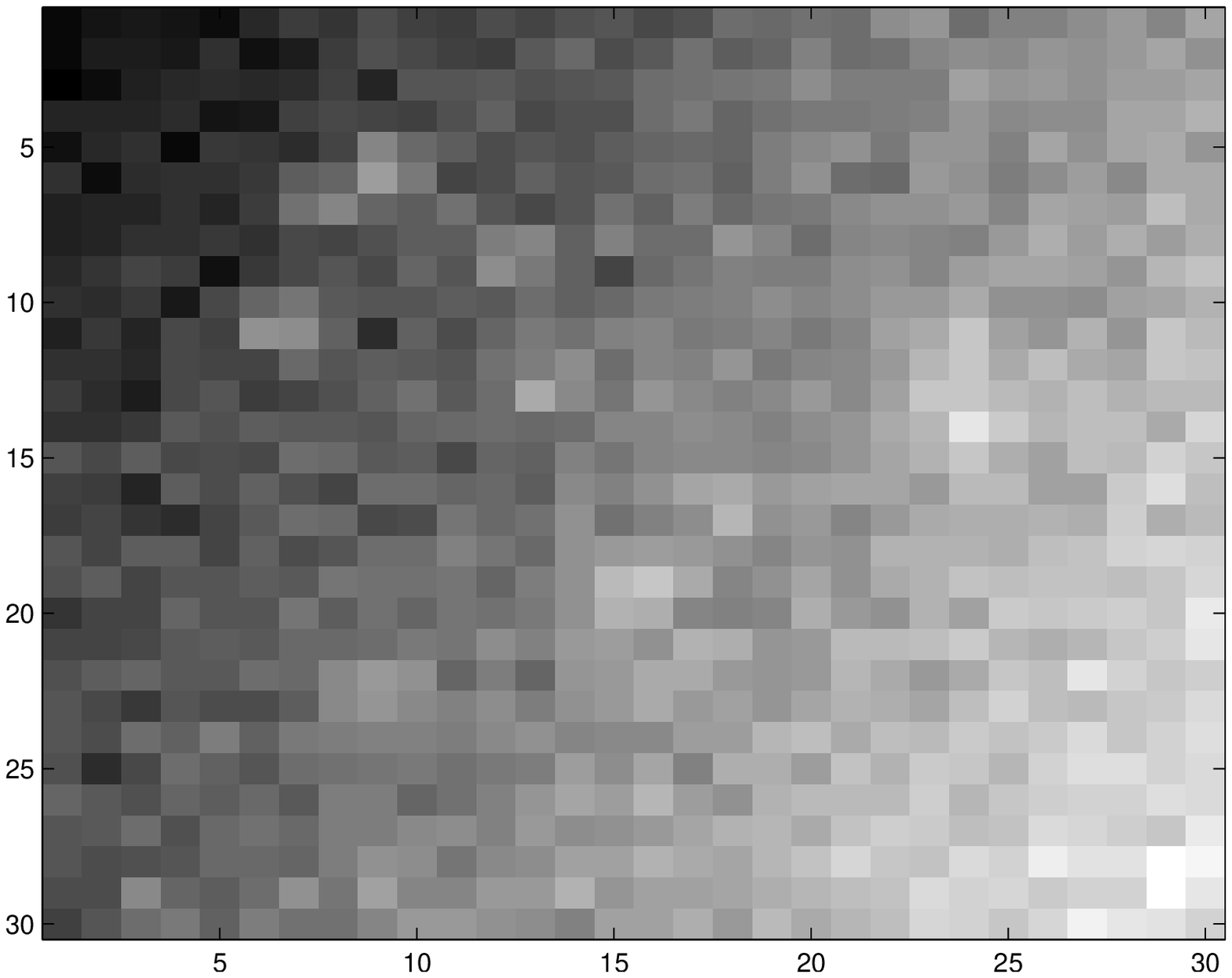}&
\includegraphics[width=0.9in, height=0.9in]{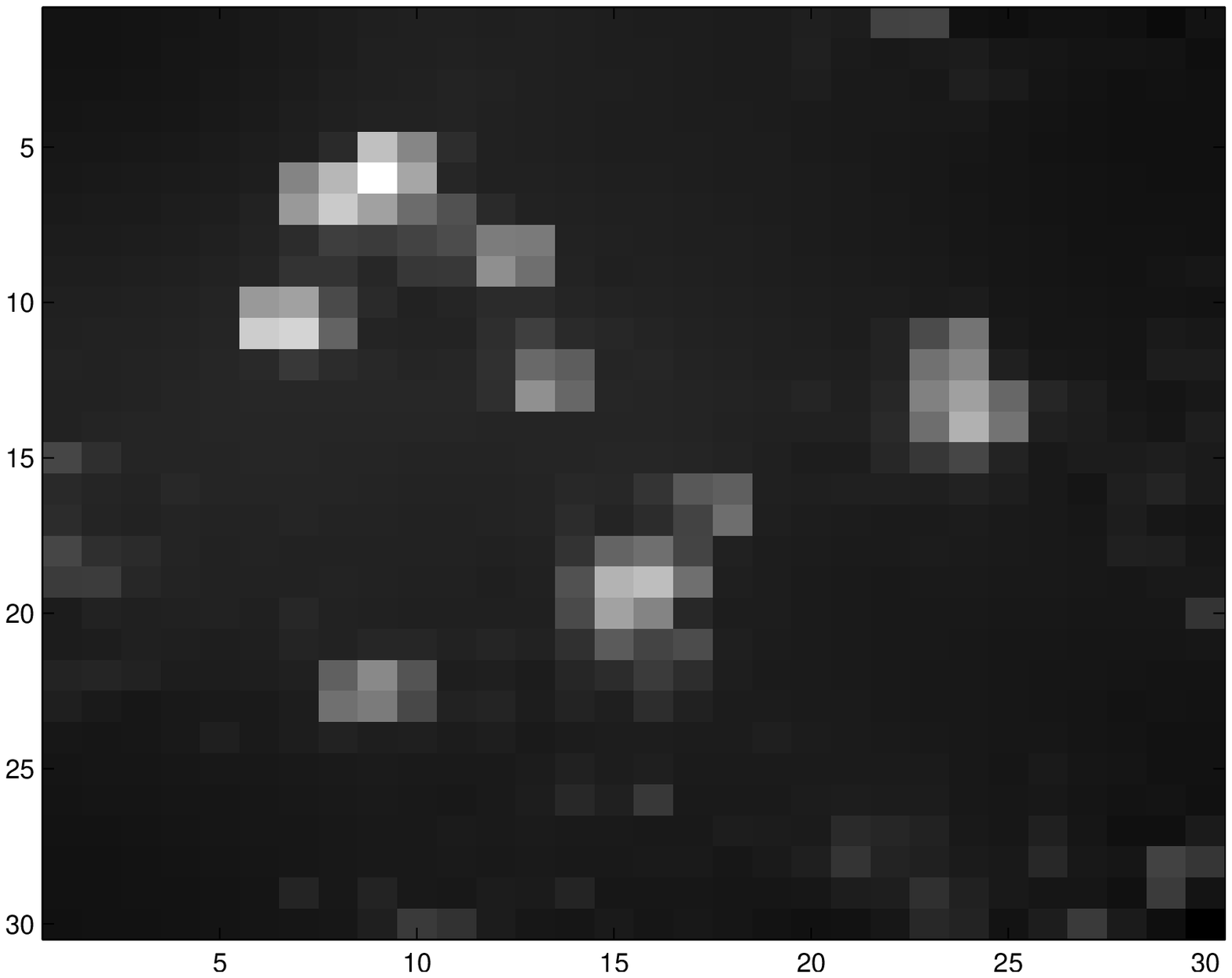}\cr
(g) Noisy image&
(h) ROF&
(i) NLM & 
(j) TV&
(g) IGMRF & 
(h) HIGMRF
\end{tabular}
\caption{(a) A simulated image containing signal from several fluorescent spots, (g) a noisy image after addition of noise resulting in signal to noise ratio in the range of $5-10$ dB  and (b, c, d, e, f, h, i, j, g, h) the denoised images after application of various de-noising algorithms}
\label{fig: Denoised images}
\end{figure*}

\subsection{Convergence Diagnostics}
\label{section: Convergence Diagnostics}
Convergence of the MCMC method is checked through the potential scale reduction factor (PSRF) diagnostic of  \cite{Gelman92:MCMCconvergence} that monitors selected outputs of the MCMC algorithm and is perhaps more accurately described as \emph{output analysis} \cite{Brooks98:MCMCconvergence,Brooks98:MCMCconvergence2}.  For $m$ scalar valued sequences of length $L$, $\{\psi^{i,n}\}_{n=0}^{L-1}$, $i=1,\ldots,m$, we define the following quantities:
\begin{eqnarray}
s_{i} &=& \frac{\sum_{n=0}^{L-1}(\psi^{i,n}-\bar{\psi}^{i})^{2}}{L-1},\quad\textrm{where }\bar{\psi}^{i}=\frac{\sum_{n=0}^{L-1}\psi^{i,n}}{L}.
\label{eq:psrfCalc}\cr
W     &=& \frac{1}{m}\sum_{i=1}^{m}s_{i},\quad B=\frac{L}{m-1}\sum_{i=0}^{m}(\bar{\psi}^{i}-\bar{\psi})^{2},\quad
\label{eq:psrfCalc_1}
\end{eqnarray}
where $\bar{\psi}=\frac{1}{m}\sum_{i=0}^{m}\bar{\psi}^{i}$. The $PSRF$ is defined to be
\[
PSRF=\left(1-\frac{1}{L}\right)+\frac{B}{LW}.
\]
The $PSRF$  is computed for all univariate parameters separately following burn-in. Convergence is consistent with $PSRF$ values below to 1.2  \cite{Gelman92:MCMCconvergence}. If this criterion is not satisfied for all parameters then it is indicative of lack of convergence and
the simulation run needs to be lengthened. 

\subsection{Evaluation of the denoising performance}
\label{section: performance evaluation}

Let ${\bf f}^{*}$ and $\tilde{\bf f}$ be the estimated image by a particular de-noising algorithm and its ground truth image respectively.

\subsubsection{Root Mean Square Error (RMSE)}
The most common and simplest metric for the image quality assessment is Root Mean Square Error (RMSE) which is computed by averaging the squared intensity differences of ${\bf f}^{*}$ and $\tilde{\bf f}$ by $RMSE({\bf f}^{*}) = \sqrt{{\bf E}( ({\bf f}^{*}-\tilde{\bf f})^{2})}$.

\subsubsection{Peak Signal-to-Noise Ratio (PSNR)}
Peak Signal-to-Noise Ratio (PSNR) is a well-known quality metric which is an extension of the RMSE by $PSNR({\bf f}^{*}) = 20\log_{10}\left( \frac{\max ({\bf f}^{*})}{RMSE({\bf f}^{*})} \right)$ where $\max({\bf f}^{*})$ is the maximum intensity of the estimated image pixels.

\subsubsection{Kullback Leibler Distance (KLD)}
Let $p_{\tilde{\bf f}}$ and $p_{{\bf f}^{*}}$ be the true distribution and the estimated distribution respectively by $KLD({\bf f}^{*}) = \sum_{i} p_{\tilde{\bf f}}(i)\log \frac{p_{\tilde{\bf f}}(i)}{p_{{\bf f}^{*}}(i)}$ where the distributions are estimated via quantification into $N_{q}$ bins. In this paper, we set $N_{q}=10$.

\subsubsection{Structural SIMilarity (SSIM)}
The other metric is structural similarity based image quality assessment which is able to explain the strong dependencies between pixels of the highly structured images \cite{Wang04imagequality}.
\begin{equation}
SSIM({\bf f}^{*}) = \frac{(2\mu_{{\bf f}^{*}}\mu_{\tilde{\bf f}}+C_{1})(2\sigma_{{\bf f}^{*}\tilde{\bf f}}+C_{2})}{(\mu_{{\bf f}^{*}}^{2}+\mu_{\tilde{\bf f}}^{2}+C_{1})(\sigma_{{\bf f}^{*}}^{2}+\sigma_{\tilde{\bf f}}^{2}+C_{2})}
\label{eq: SSIM}
\end{equation}
where $\mu_{{\bf f}^{*}} = \sum_{i}f^{*}_{i}/N$, $\mu_{\tilde{\bf f}} = \sum_{i}\tilde{f_{i}}/N$, $\sigma_{{\bf f}^{*}} =\left[\sum_{i}(f^{*}_{i}-\mu_{{\bf f}^{*}})^{2}/N\right]^{1/2}$, $\sigma_{\tilde{\bf f}} = \left[\sum_{i}(\tilde{f_{i}}-\mu_{\tilde{\bf f}})^{2}/N\right]^{1/2}$, and $\sigma_{{\bf f}^{*}\tilde{\bf f}} = \sum_{i}(f^{*}_{i}-\mu_{{\bf f}^{*}})(\tilde{f_{i}}-\mu_{\tilde{\bf f}})/N$. In this paper, for simplicity we use $C_{1}=C_{2}=0$ which is commonly defined as "universal quality index (UQI)" although this produces unstable results when the denominator of Eq. (\ref{eq: SSIM}) is very close to zero \cite{Wang02:UQI}. 

\section{Application to synthetic and real single molecule data}
\label{section: Simulated Results}
\subsection{Synthetic images}

\begin{figure}[h!]
\centering
\begin{tabular}{cc}
\includegraphics[width = 3in, height=2.6in]{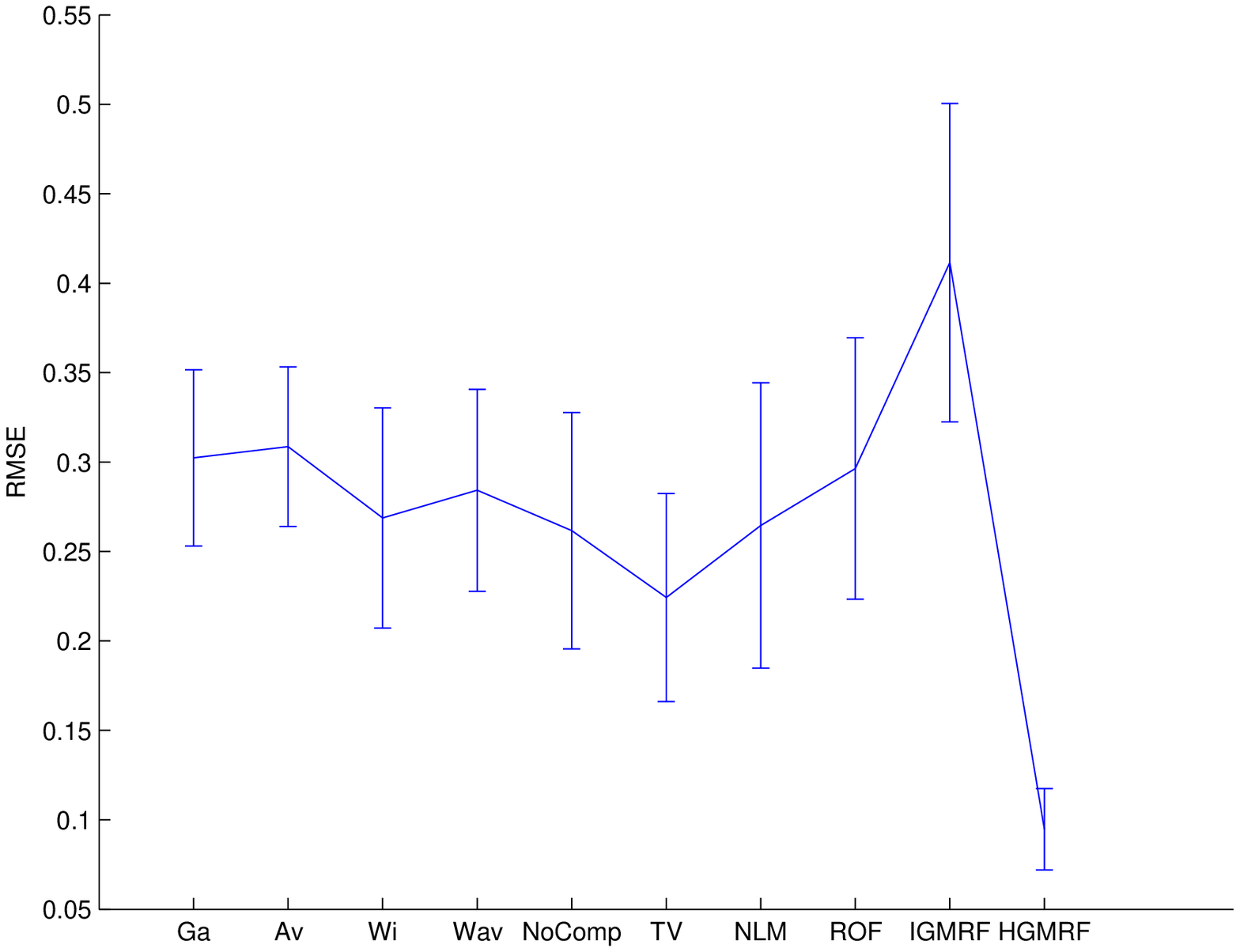}&
\includegraphics[width = 3in, height=2.6in]{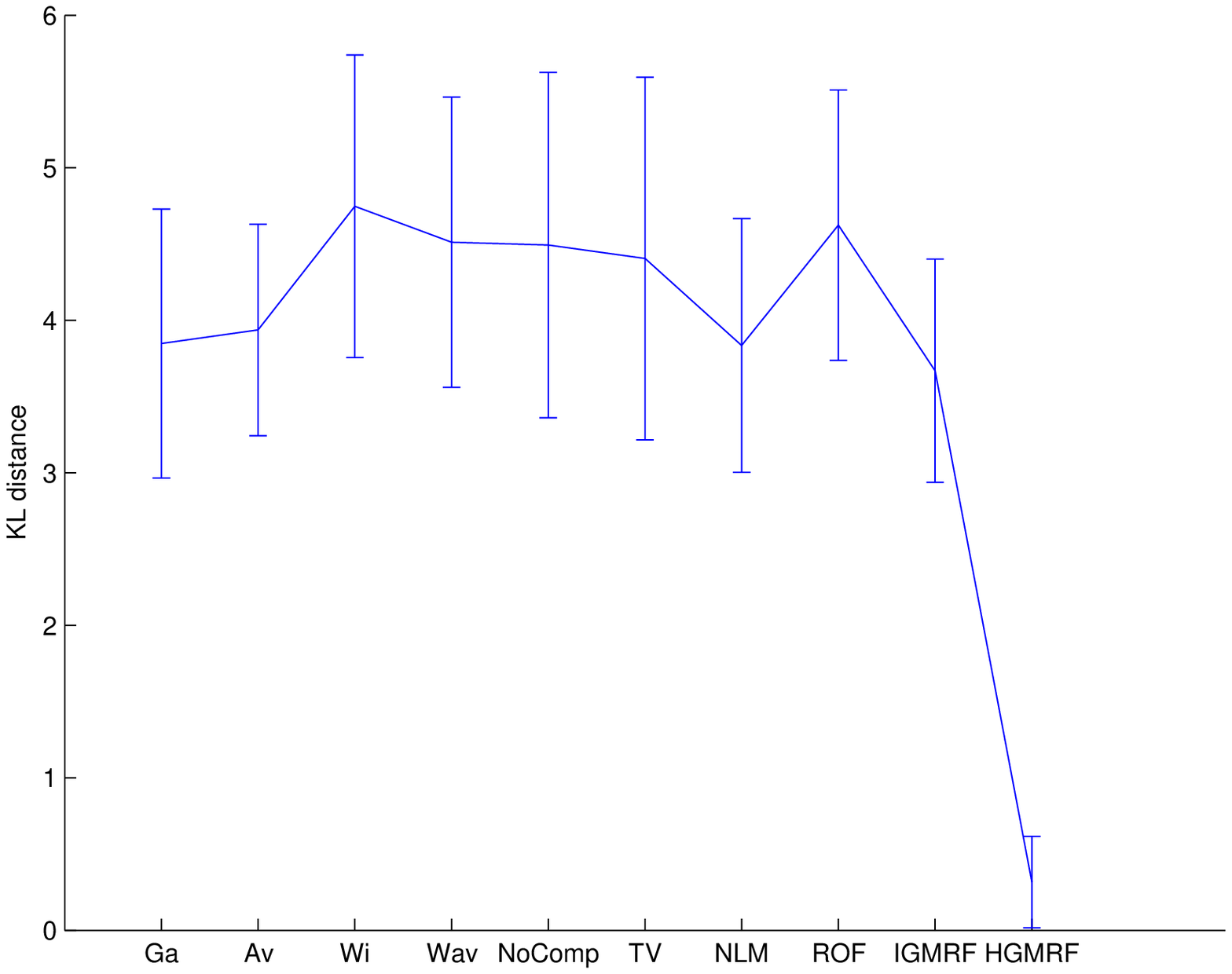}\cr
(a) RMSE&
(b) KLD\cr
\includegraphics[width = 3in, height=2.6in]{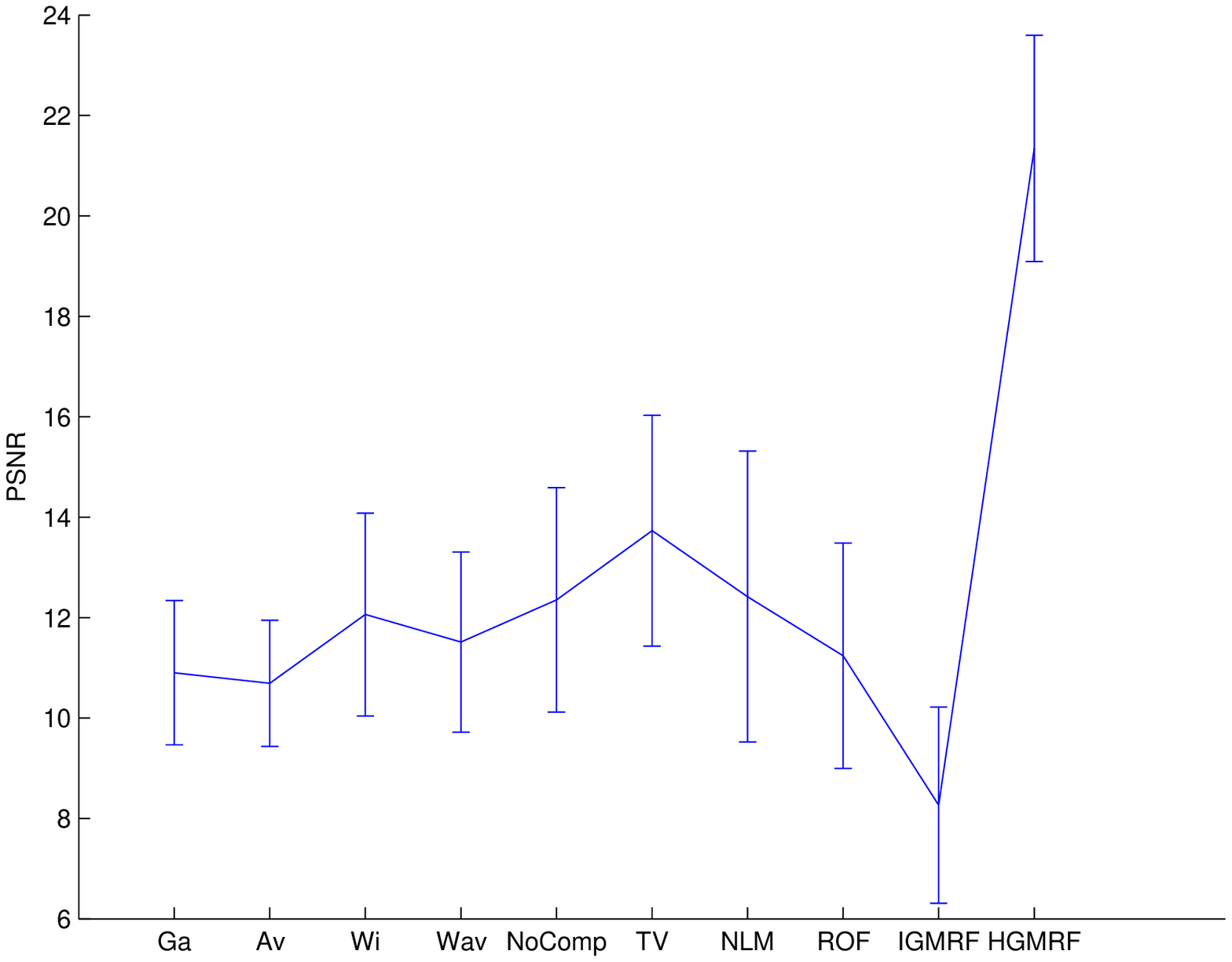}&
\includegraphics[width = 3in, height=2.6in]{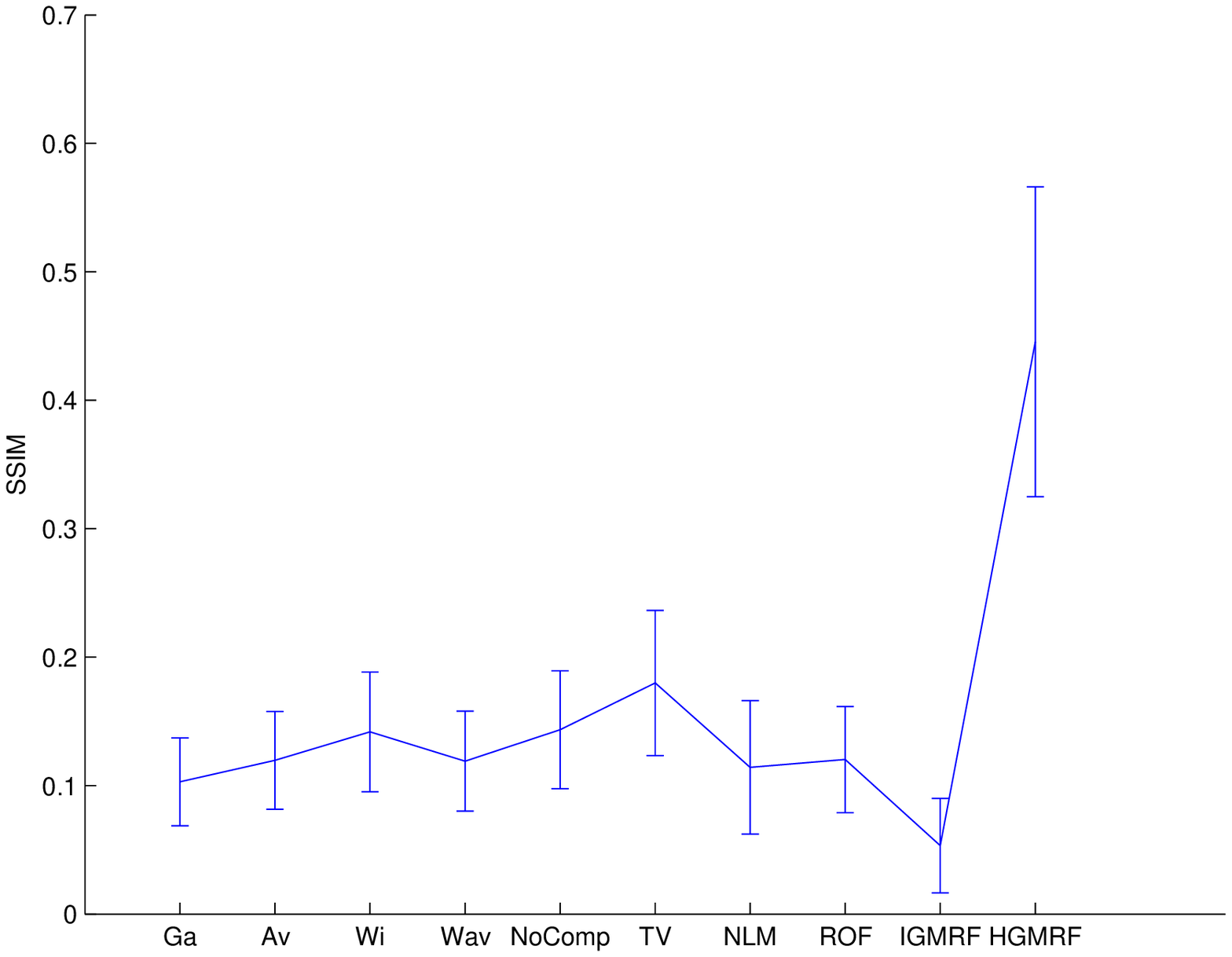}\cr
(c) PSNR&
(d) SSIM
\end{tabular}
\caption{Mean and standard deviation of Root Mean Square Error (RMSE), Kullback Leibler Distance (KLD), Peak Signal-to-Noise Ratio (PSNR) and Structural SIMilarity (SSIM) for 50 random images.}
\label{fig: Root Mean Square Error}
\end{figure}

Initially we sought to quantitatively compare the performance of the various algorithms using synthetic data. Synthetic data consisted of 50 randomly generated ground-truth images of a $30 \times 30$ pixel region with a small number of single molecule signals, to which varying levels of noise were added (resulting in signal to noise ratios in the range of 5-10 dB for different noise levels). The various parameters used in generation of this data are given in Table I. Next, various filtering algorithms were applied to the data. Figure \ref{fig: Denoised images} shows an example of simulated data (Fig. \ref{fig: Denoised images} a and g) and the de-noised images obtained through different algorithms: Gaussian filter (Ga), Average filter (Av), Wiener filter (Wi), Wavelet filter (Wav) \cite{Donoho93:nonlinearwavelet}, Log Garbor Wavelet (LGW) based filter \cite{Kovesi_phasepreserving}, Total Variation approach by Rubin et al. (ROF) \cite{Rudin92:TV}, Non-Local Mean filter (NLM) \cite{Buades05:anon-local}, Total Variation by Split Bregman Method (TV) \cite{Goldstein09:Denoising}, IGMRF, and HIGMRF of Fig. \ref{fig: Denoised images}. By eye it appeared that the HIGMRF filtered images came closest to matching the underlying ground-truth. Analyzing the root mean square error (RMSE), Kullback Leibler Distance (KLD), Peak Signal-to-Noise Ratio (PSNR) and Structural SIMilarity (SSIM) between the intensities of the ground truth relative to recovered intensities in the filtered images (Fig. \ref{fig: Root Mean Square Error}) confirmed this supposition, showing that intensities recovered after HIGMRF were much closer to the truth than the other methods.
\begin{table}[h!]
\caption{Systematic parameters for synthetic images}
\label{table: Systematic parameters for synthetic images}
\centering
\begin{tabular}{||c|r||c|r||}\hline\hline
parameter & value & parameter & value\cr\hline
$h$& $0.1$ & $T$& $100$\cr
$\alpha_{l}$& $1$ & $\beta_{l}$ & $10$\cr
$\alpha_{f}$& $10$ & $\beta_{f}$ & $0.01$\cr
\hline\hline
\end{tabular}
\end{table}

\subsection{Single molecule fluorescence images}

The performance of the algorithms was then tested with real single molecule fluorescence data, obtained on living human T-cells, a cell of the immune response. To create a suitable sample, the immunoglobulin protein CD86 was genetically introduced into human T-cells so that it was present at the cell membrane at low concentrations, as described in Dunne \textit{et al.}\cite{Dunne09:SMF}. These molecules were labelled with an organic dye-labelled antibody fragment specific to CD86 (Alexa Fluor 488 labelled Fab molecule, derived from the anti-CD86 antibody BU63, see James \textit{et al.}\cite{James07:SingleMolecule} for details), and imaged using Total Internal Reflection Fluorescence Microscopy, as described previously \cite{Dunne09:SMF}.

\begin{figure}[h!]
\centering
\begin{tabular}{c|ccccc}
\includegraphics[width=0.9in, height=0.9in]{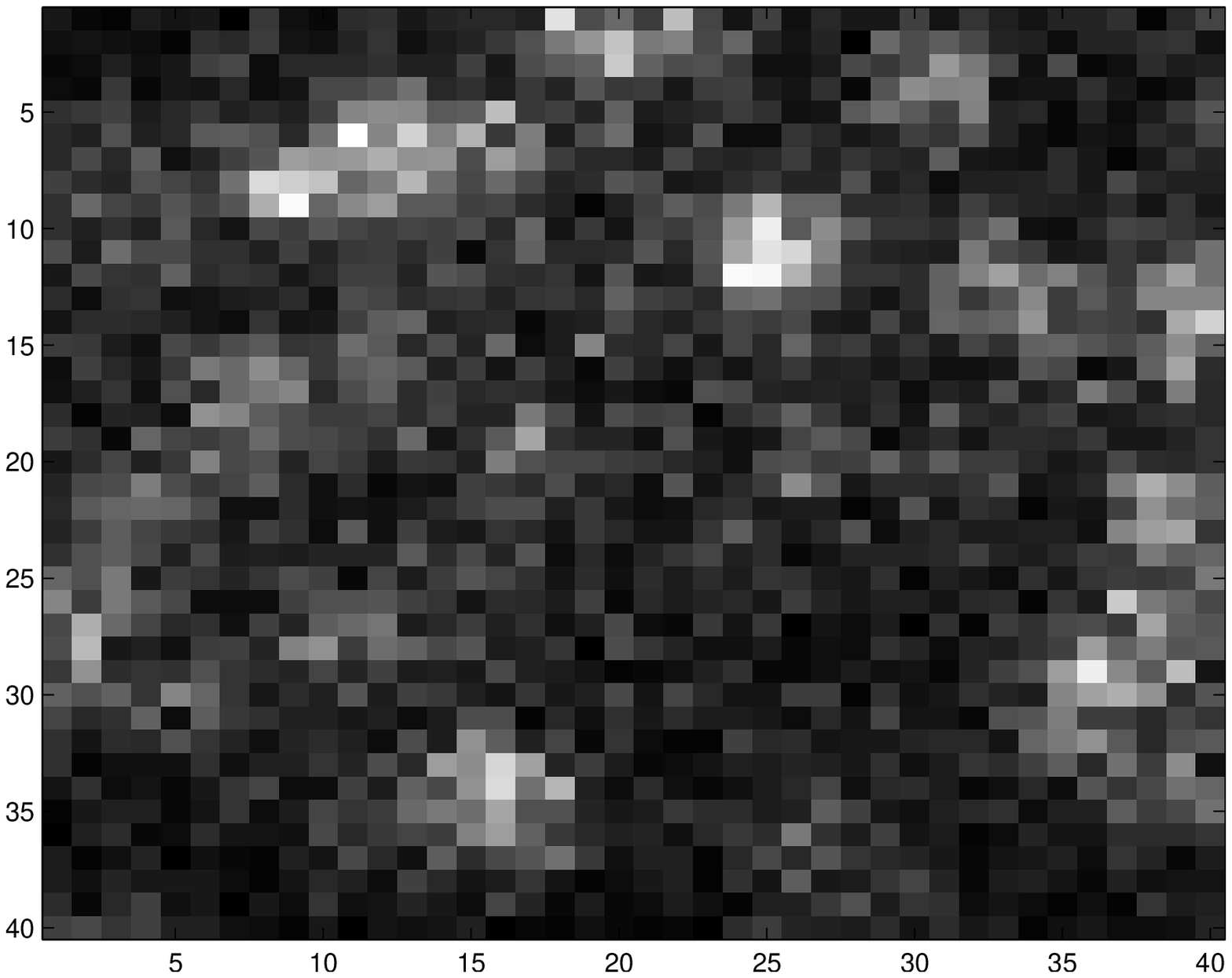}&
\includegraphics[width=0.9in, height=0.9in]{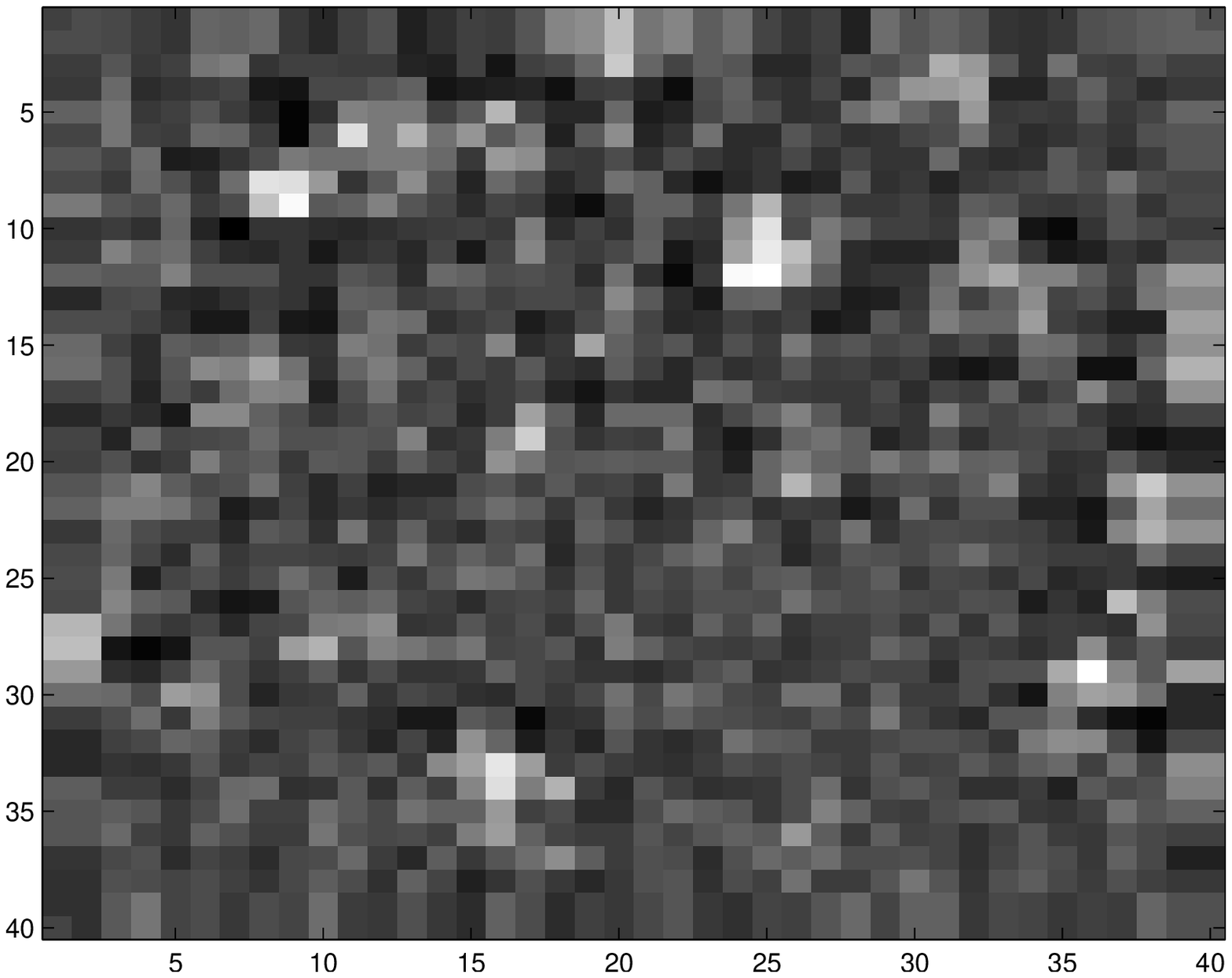}&
\includegraphics[width=0.9in, height=0.9in]{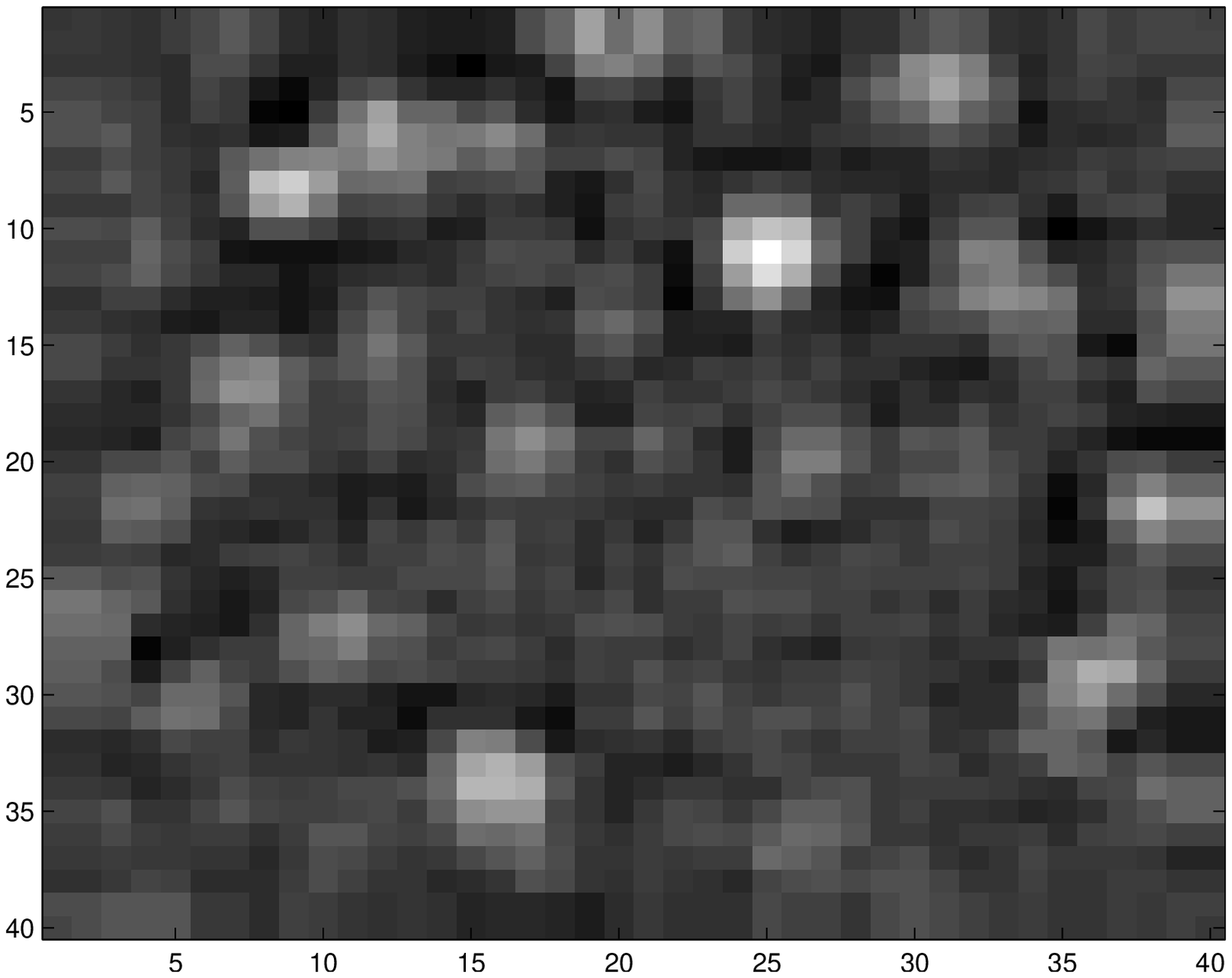}&
\includegraphics[width=0.9in, height=0.9in]{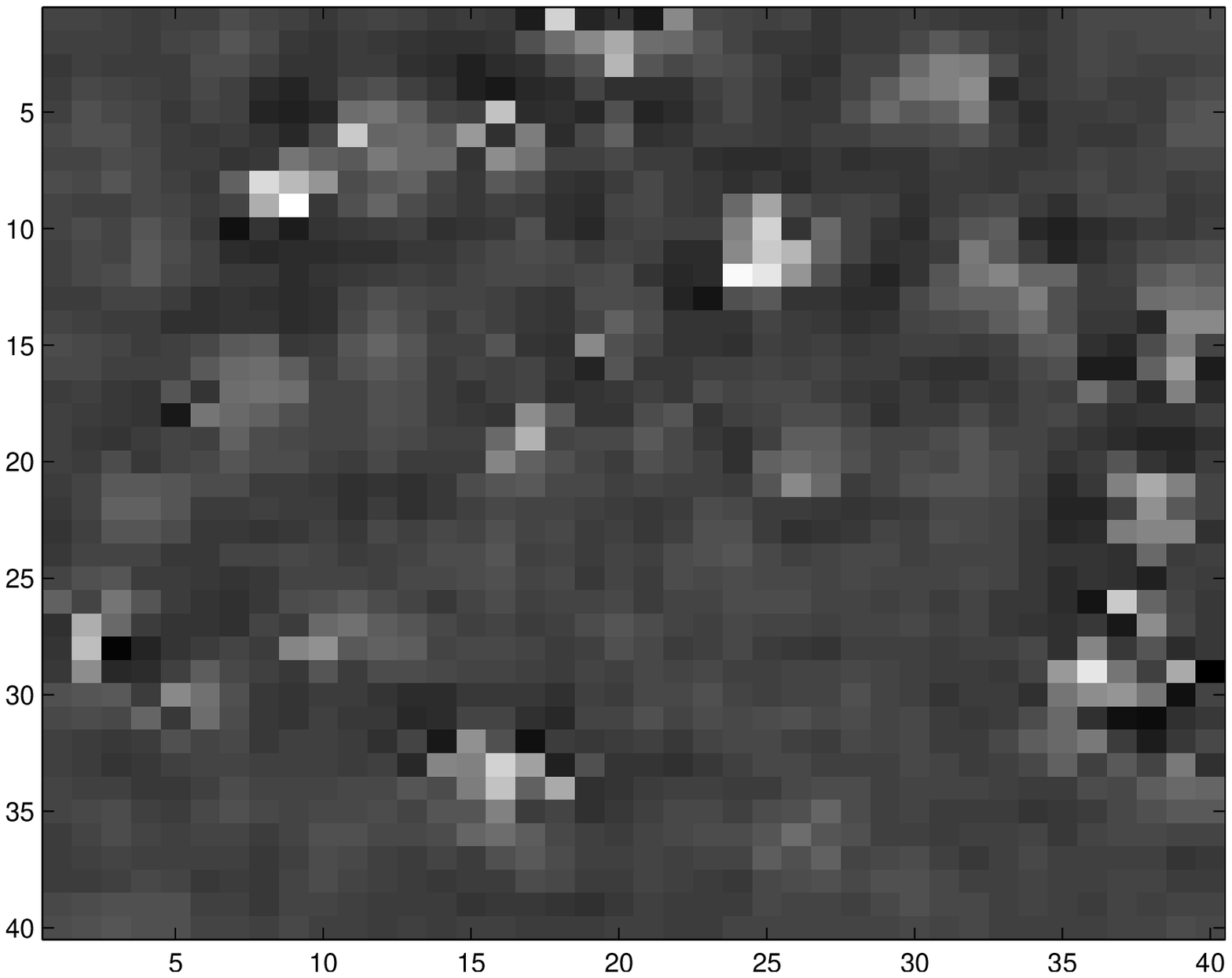}&
\includegraphics[width=0.9in, height=0.9in]{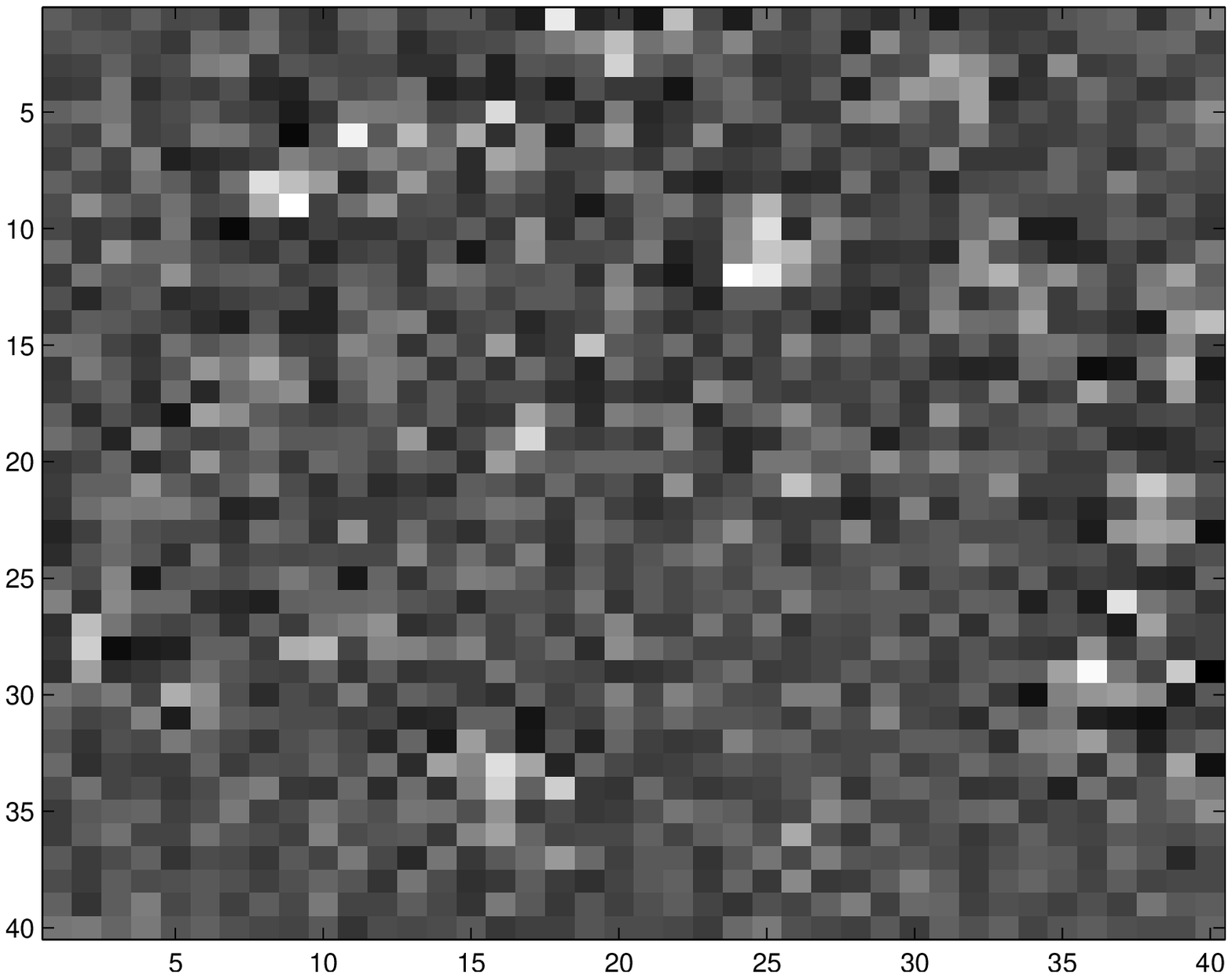}&
\includegraphics[width=0.9in, height=0.9in]{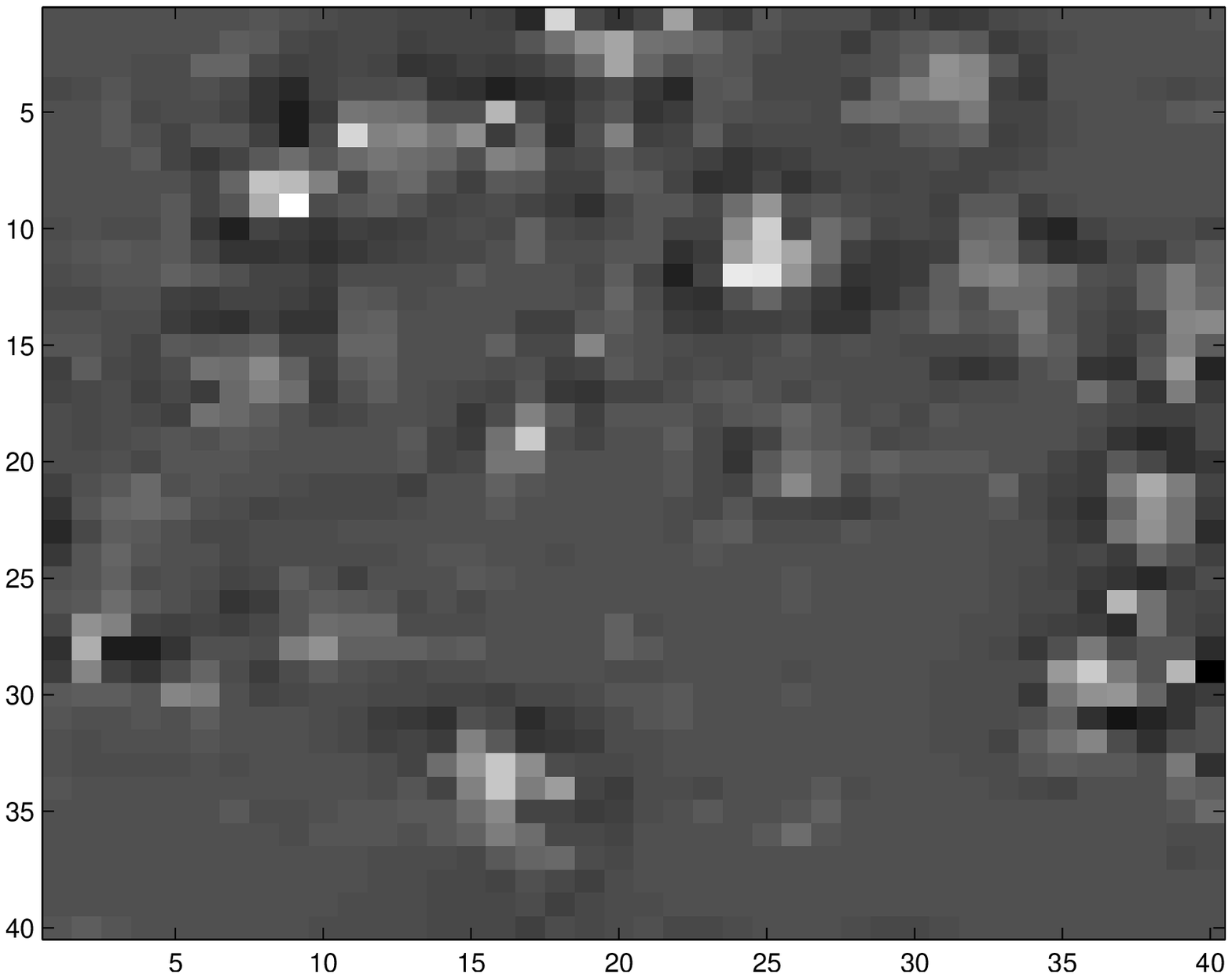}\cr
(a) Observation&
(b) Gaussian& 
(c) Average&
(d) Wiener& 
(e) Wavelet&
(f) LGW\cr
&
(Ga) & 
(Av)&
(Wi)& 
(Wav)&
\cr
&
\includegraphics[width=0.9in, height=0.9in]{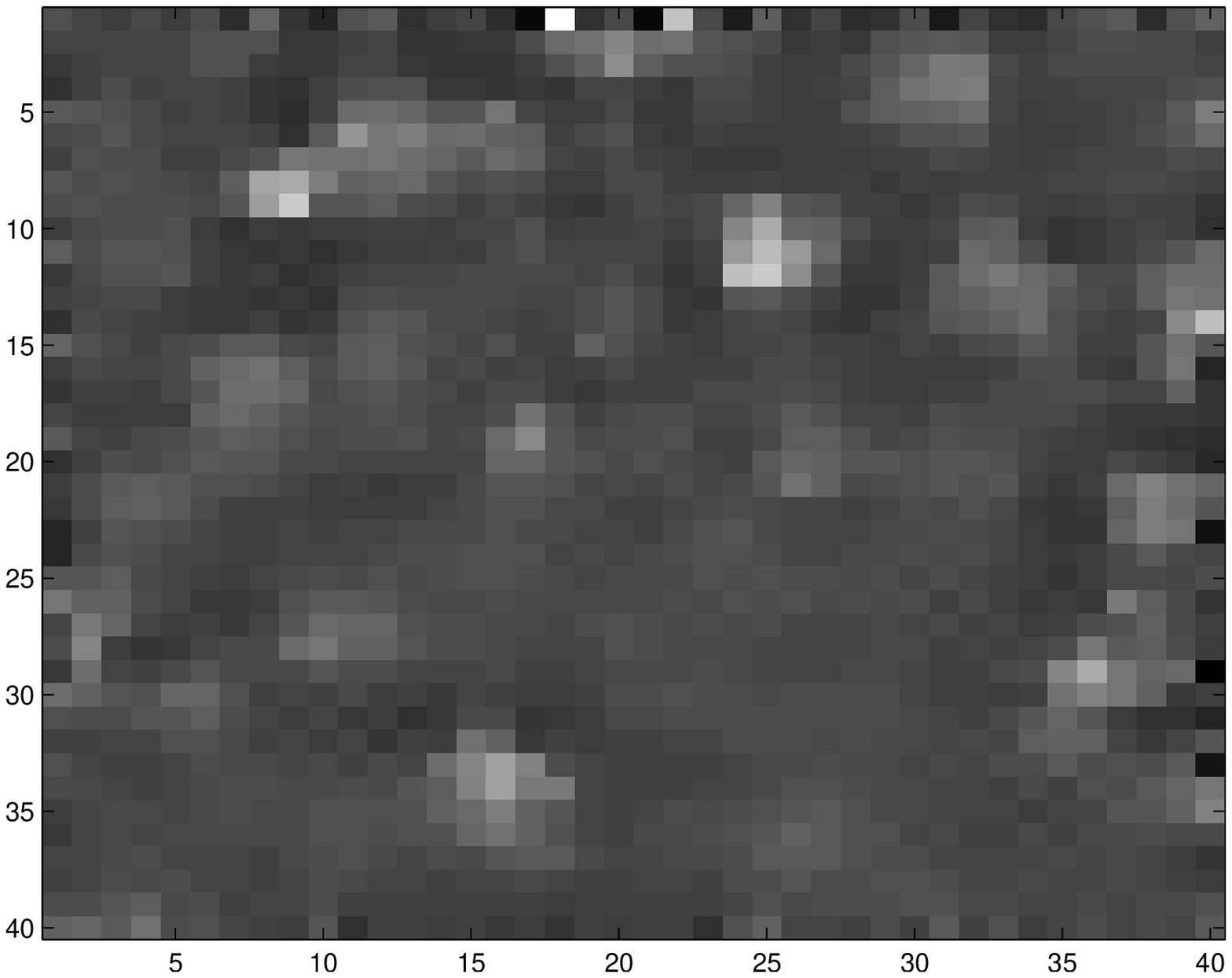}&
\includegraphics[width=0.9in, height=0.9in]{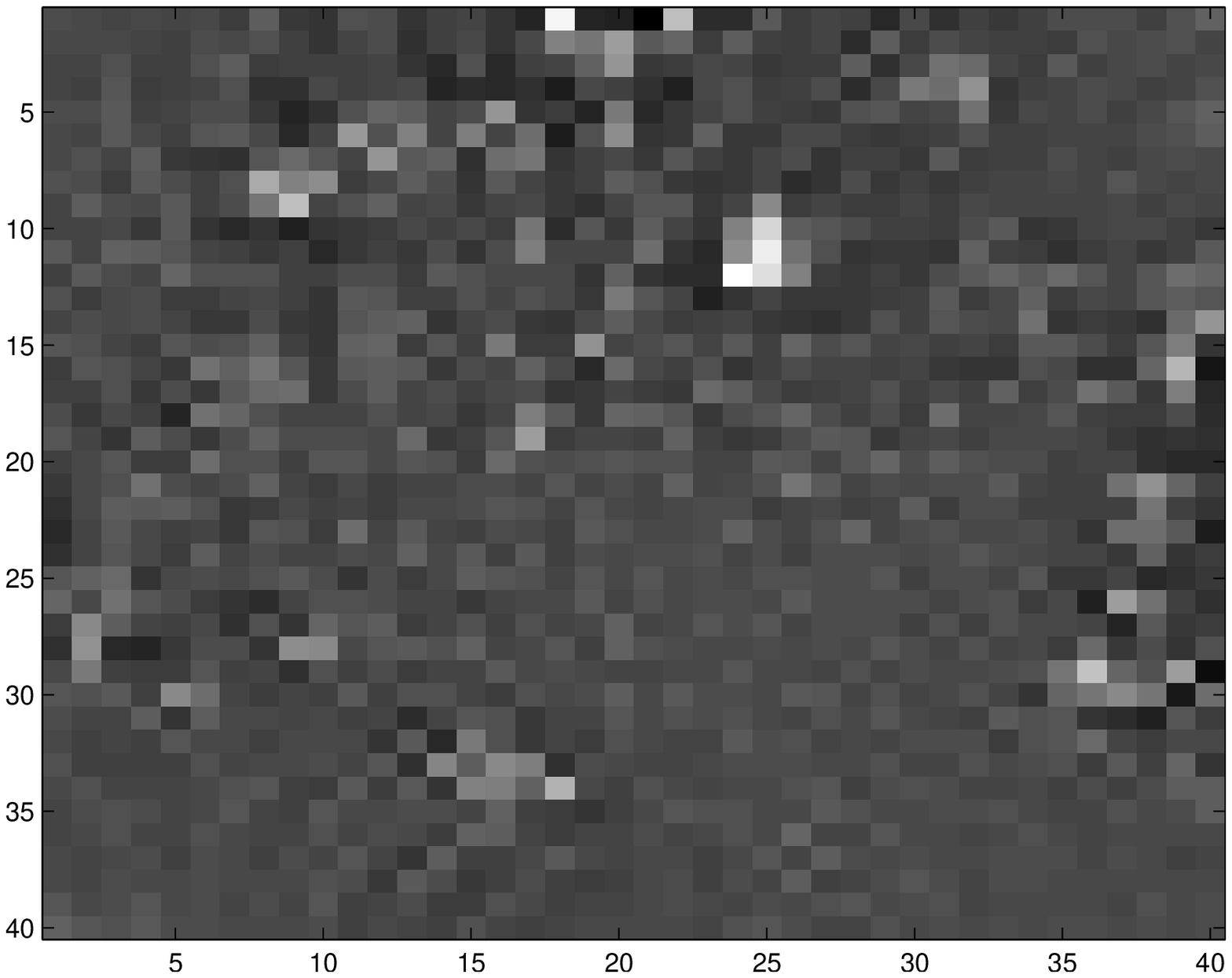}&
\includegraphics[width=0.9in, height=0.9in]{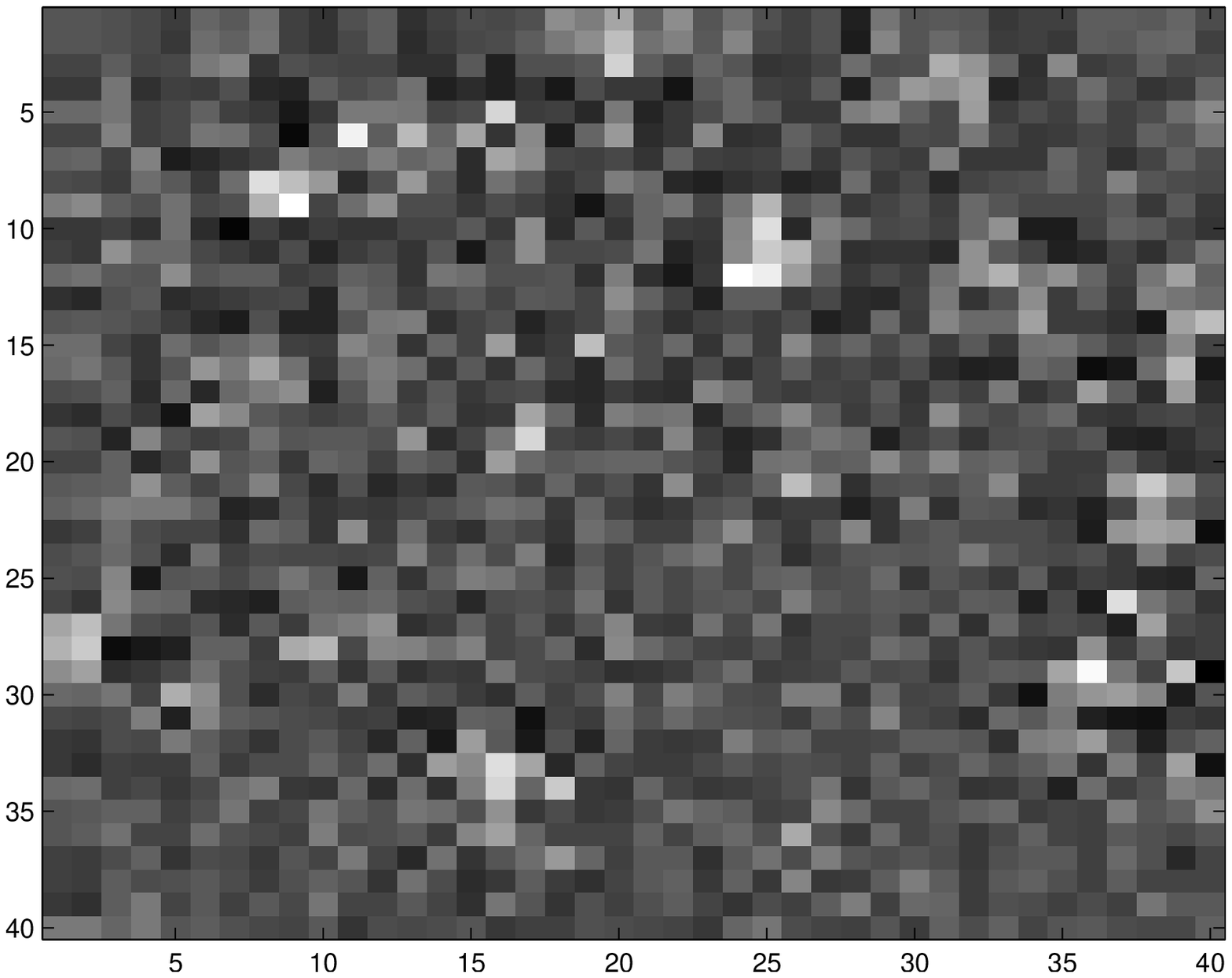}&
\includegraphics[width=0.9in, height=0.9in]{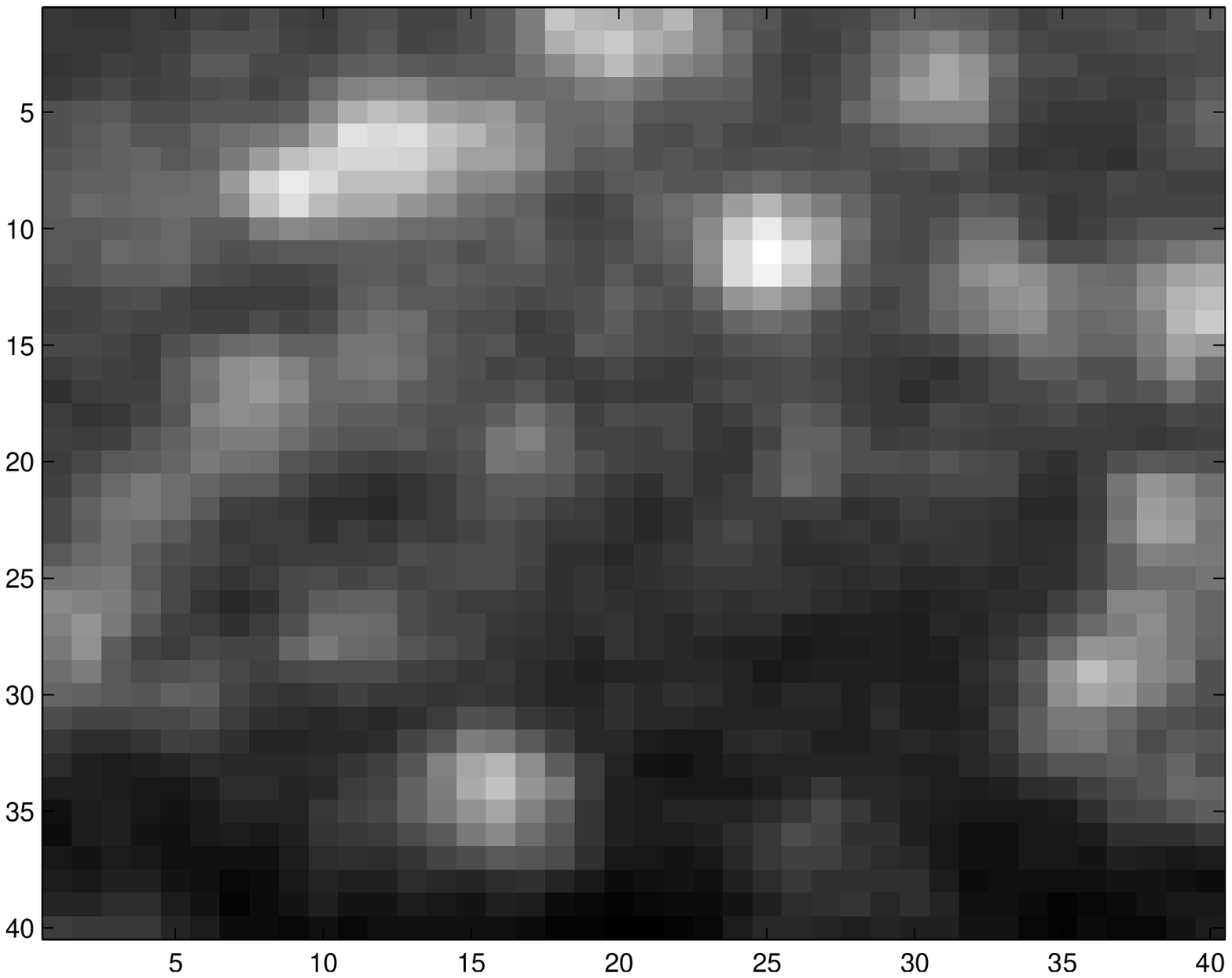}&
\includegraphics[width=0.9in, height=0.9in]{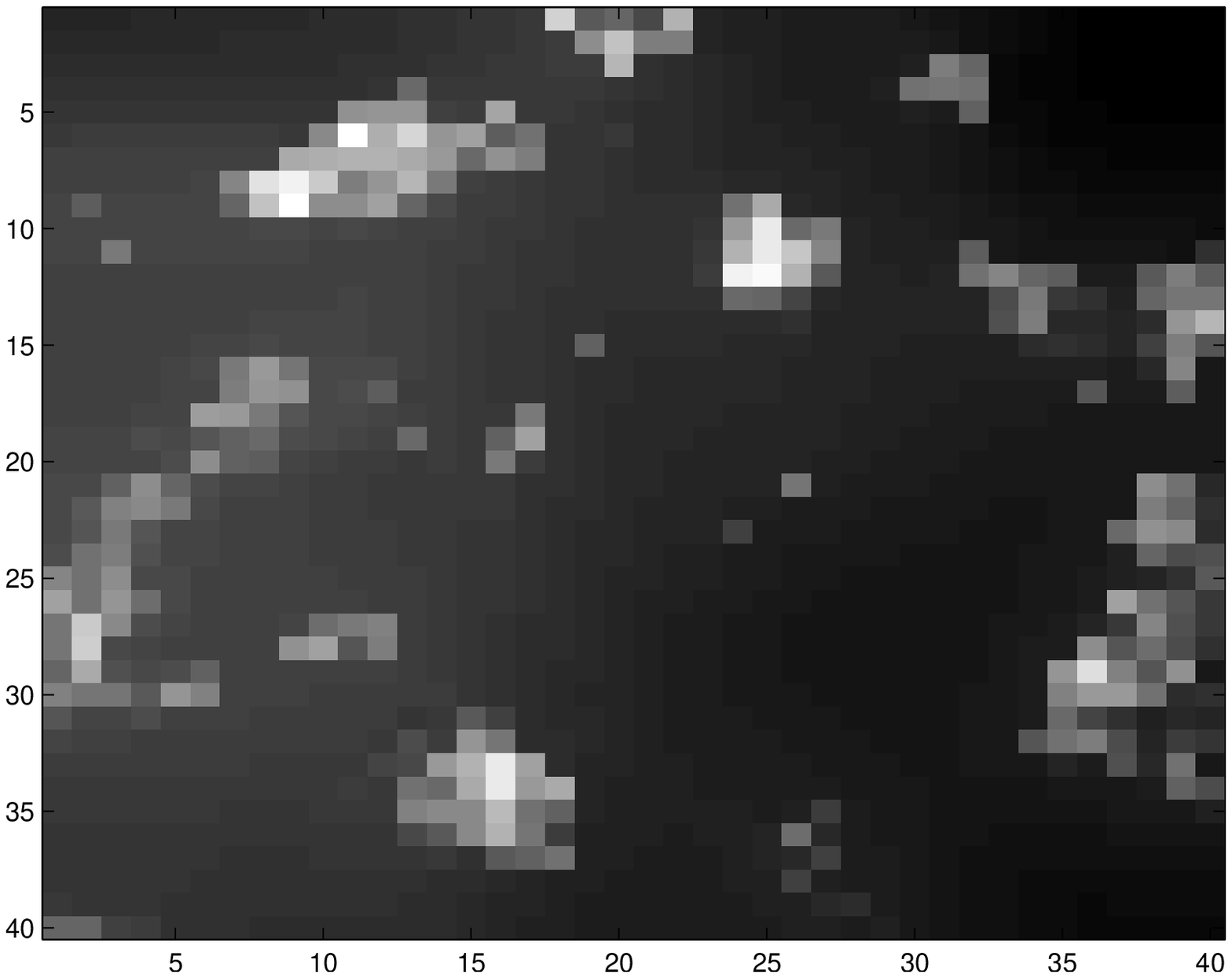}\cr
& 
(g) ROF&
(h) NLM& 
(i) TV&
(j) IGMRF & 
(k) HIGMRF
\end{tabular}
\caption{(a) An experimental and noisy single molecule fluorescence image containing signal (the 20th frame of CD86 diffusing within the membranes of live T-cells) and (b-k) the restored images after application of various de-noising algorithms.}
\label{fig: Results for the experimental data}
\end{figure}

\begin{figure}[h!]
\centering
\begin{tabular}{cc}
\includegraphics[scale=0.43]{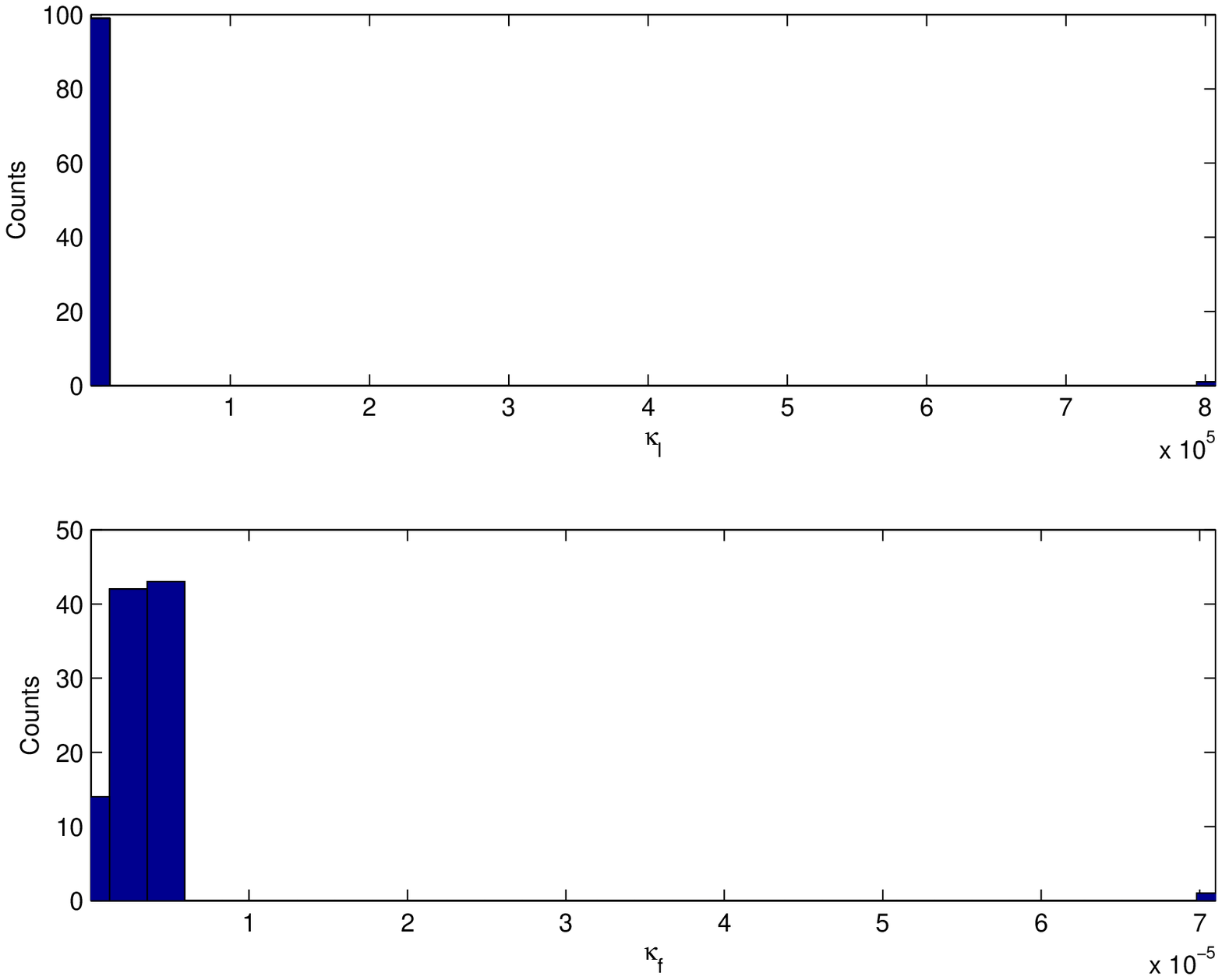}&
\includegraphics[scale=0.43]{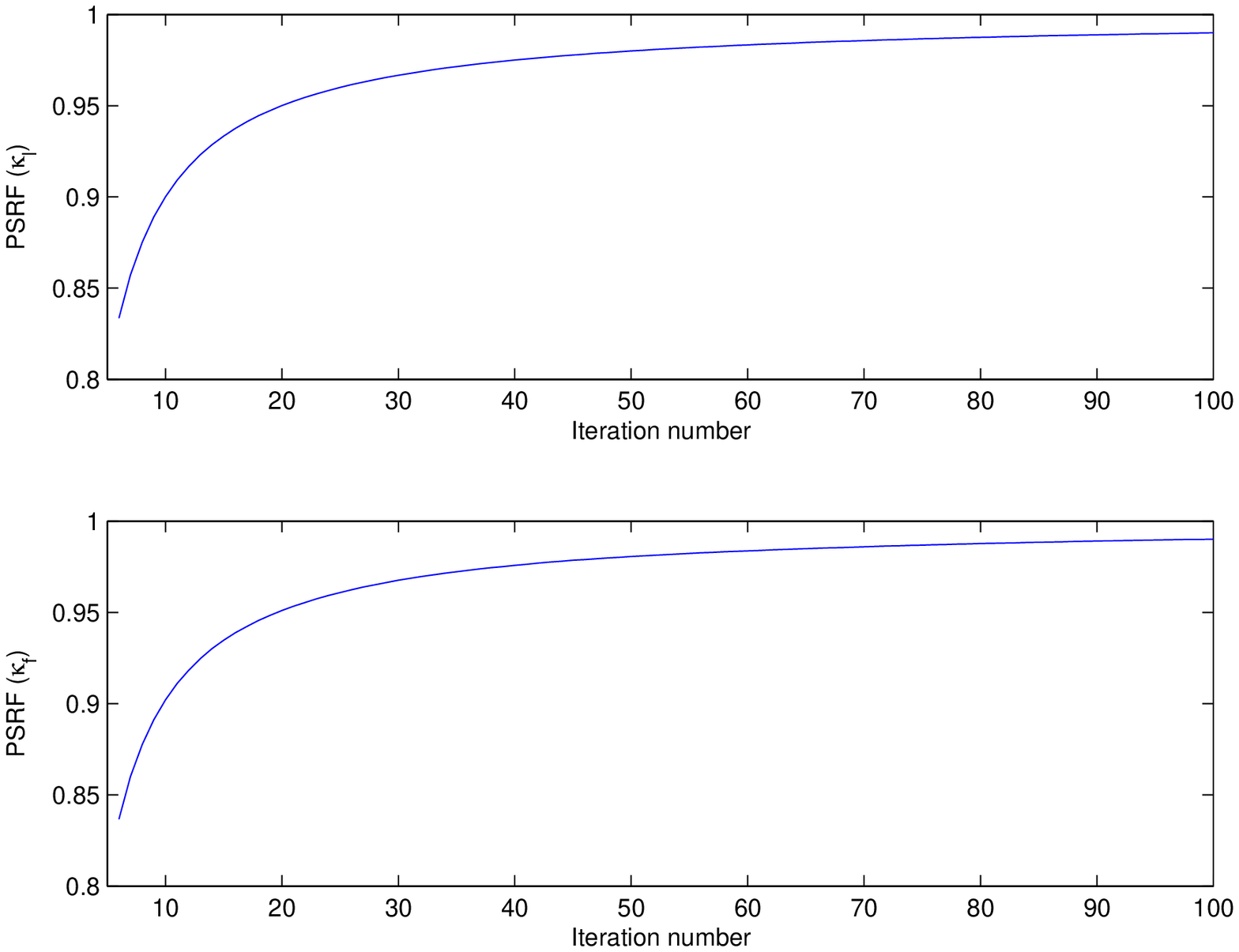}\cr
(a) Histograms of sampled values of $\theta$&
(b) The convergence of $\theta$
\end{tabular}
\caption{Samples of $\theta$: (a) histogram of samples $\theta$ generated over $p(\theta|{\bf y})$ and (b) the convergence checking of $\theta$ by the PSRF method.}
\label{fig: Samples of theta}
\end{figure}

For this real experimental image, the performances of ten different de-noising algorithms as shown in synthetic dataset were compared for one frame (Fig. \ref{fig: Results for the experimental data} a). The results from these filtering steps are shown in Fig. \ref{fig: Results for the experimental data} (b)-(k). The figures show only chopped ared ($40\times 40$ size) to clearly see the denoising effect. Compared to other approaches the HIGMRF preserves the spot signals of interest while it suppresses the global and local noises effectively. The convergence of our Gibbs sampler during IGMRF and HIGMRF filtering of this data was checked using the PSRF diagnostic method described in Section \ref{section: Convergence Diagnostics}. The PSRF was found to converge to a value below 1.2 for all components of $\theta$ (Fig. \ref{fig: Samples of theta}).

\section{Discussion}

Several issues for the current proposed algorithm remain to be resolved and discussed. First of all, inferring ${\bf e}$ is a key step for the proposed algorithm. Basically, ${\bf e}$ is regarded as a set of clustered labels given continuous values ${\bf f}$. Therefore, we can apply any clustering algorithms with two clusters. For instance, we can partition (cluster) ${\bf f}$ into $K$ different clusters by using $K$-means algorithms to infer ${\bf e}$ with $K=2$. However, it is rather difficult to build locality based clusters and it might infringe the properties of Gibbs sampling. The best way to reconstruct ${\bf e}$ is to generate them in a Gibbs sampling scheme (that is, single-spin updates which is called 'heat method' in the statistical physics). However, it takes a lot of time to reach the convergence since the dimension of ${\bf e}$ is too large to obtain sufficiently good results in a practical time. (As we know, Gibbs sampling requires a myriad of samples to restore joint posterior when variables (dimensions) are highly dependent.) Worse, the variables of ${\bf e}$ are not continuous variables but binary variables, which are often rather difficult to infer. We can apply the Swendsen-Wang model which is known as relatively fast sampling algorithm for Ising model by adding auxiliary variable for bonds. However, inferring ${\bf e}$ via Monte carlo sampling even with such a fast Swendsen-Wang model is still problematic since it brings a lot of time consuming tasks for the convergence. Therefore, we adopt a simple threshold algorithm in order to reduce time complexity although the threshold based algorithm has the sensitivity problem.  The threshold based algorithm can be regarded as a sort of a simple clustering algorithm. Also, it works as if the proposal function is a Dirac delta function in Gibbs sampling. From this point of view, the simple threshold approach is practically robust and useful. In this current version, we set a very low value by $0.1$ in order to allow relatively dim spot regions to be classified. The threshold may be estimated in the computation as well, although the fixed value worked well in the examples described here. As a further work, one might consider assuming $h$ unknown and  try to infer its value in the Gibb's sampler scheme. In addition, we may be able to jointly generate samples of ${\bf e}$ and ${\bf f}$ on by embedding Bernoulli distribution into the system. This joint estimation will improve our current algorithm a lot since it is known that such a highly dependent parameters can be improved by blocked Gibbs sampling technique. 

In addition, although most of the hyper-parameters in the model ---  $\alpha_{l}$ and $\beta_{l}$ for $\kappa_{l}$ --- can be assigned vague priors, an informative prior on $\alpha_{f}$ and $\beta_{f}$ is necessary. These are the hyper-parameters for the precision $\kappa_f$ of the (H)IGMRF  and the solution is sensitive to their value.

\section{Conclusion}

In this paper, a hierarchical  model has been proposed to de-noise single molecule fluorescence images, in which a Gaussian Markov random field (GMRF) is used as a well defined prior of a latent Gaussian model. A variant of  the intrinsic GMRF, called the heterogeneous intrinsic GMRF (HIGMRF), was introduced in order to maintain the attractive properties and computational advantages of the intrinsic GMRF while addressing some of its weaknesses, such as blurring and stationary.  These are obtained by assigning non-stationary interaction weights to pairs of pixels of the image. The weights and hence precision matrix (inverse of covariance matrix) for the HIGMRF are updated sequentially in a Gibbs framework. It has been demonstrated that the use of the HIGMRF reduces global and local noise from images effectively.

\bibliographystyle{plain}
\bibliography{myBib}

\end{document}